\documentclass{article}

\usepackage{microtype}
\usepackage{graphicx}
\usepackage{subcaption}
\usepackage{booktabs} 
\usepackage{hyperref}

\usepackage[preprint]{icml2026}

\usepackage{mathtools}
\usepackage{amsthm}
\usepackage{enumitem}
\usepackage{makecell}

\usepackage[utf8]{inputenc} 
\usepackage[T1]{fontenc}    
\usepackage{url}            
\usepackage{amsfonts}       
\usepackage{nicefrac}       
\usepackage[table]{xcolor}\usepackage{float}
\usepackage{gensymb}
\usepackage{amsmath,amssymb}
\usepackage{multirow}
\usepackage{svg}
\usepackage{wrapfig}
\usepackage{import}
\usepackage{todonotes}

\renewcommand{\arraystretch}{1.4}
\newcolumntype{C}{>{\centering\arraybackslash}m{3.5cm}}
\newcolumntype{L}{>{\raggedright\arraybackslash}m{3.5cm}}
\newcommand{\hl}[1]{\cellcolor{gray!15}#1}

\usepackage[capitalize,noabbrev]{cleveref}

\icmltitlerunning{Resolving Extreme Data Scarcity in Groundwater Heat Transport Applications}

\begin{document}

\twocolumn[
  \icmltitle{Resolving Extreme Data Scarcity\\by Explicit Physics Integration:\\An Application to Groundwater Heat Transport}


  \begin{icmlauthorlist}
    \icmlauthor{Julia Pelzer}{yyy}
    \icmlauthor{Corn\'{e} Verburg}{comp}
    \icmlauthor{Alexander Heinlein}{comp}
    \icmlauthor{Miriam Schulte}{yyy}
  \end{icmlauthorlist}

  \icmlaffiliation{yyy}{Institute for Parallel and Distributed Systems, University of Stuttgart, Stuttgart, Germany}
  \icmlaffiliation{comp}{Delft Institute of Applied Mathematics, Delft University of Technology (TU Delft), Delft, the Netherlands}

  \icmlcorrespondingauthor{Julia Pelzer}{julia.pelzer@ipvs.uni-stuttgart.de}

  \icmlkeywords{Machine Learning, ICML}

  \vskip 0.3in
]

\printAffiliationsAndNotice{} 
\begin{abstract}

Real-world flow applications in complex scientific and engineering domains, such as geosciences, challenge classical simulation methods due to large spatial domains, high spatio-temporal resolution requirements, and potentially strong material heterogeneities that lead to ill-conditioning and long runtimes. While machine learning–based surrogate models can reduce computational cost, they typically rely on large training datasets that are often unavailable in practice.
To address data-scarce settings, we revisit the structure of advection–diffusion problems and decompose them into multiscale processes of locally and globally dominated components, separating spatially localized interactions and long-range effects. We propose a Local–Global Convolutional Neural Network (LGCNN) that combines a lightweight numerical model for global transport with two convolutional neural networks addressing processes of a more local nature.
We demonstrate the performance of our method on city-scale geothermal heat pump interaction modeling and show that, even when trained on fewer than five simulations, LGCNN generalizes to arbitrarily larger domains, and can be successfully transferred to real subsurface parameter maps from the Munich region, Germany.
\end{abstract}

\section{Introduction}\label{sec:intro}
Many learning problems in scientific and engineering domains
\cite{sharma2021machine, sarker2021machine, angra2017machine, Jhaveri2022A} are governed by underlying physical processes. Yet they challenge classical numerical simulations due to large spatial domains, high resolution requirements, and potential ill-conditioning. In addition, many applications require repeated evaluations, for example for uncertainty quantification, design optimization, and real-time control. From a machine learning (ML) perspective, such problems are attractive targets for surrogate modeling, as trained models can dramatically reduce inference cost, even if offline training cost is substantial.

However, in real-world advection-diffusion problems, standard ML approaches face two central difficulties \cite{dietrich25}. First, training data are often scarce: measurements are indirect or sparse, and generating labeled data via high-fidelity simulations is computationally expensive. Second, the underlying dynamics exhibit multiscale behavior with both short-range local interactions and long-range global dependencies, making it difficult for data-driven models to generalize in data-scarce regimes.
Convolutional architectures such as UNets \cite{CNN15} effectively model local spatial patterns~\cite{thuerey_airfoils, Jhaveri2022A}, but representing non-local interactions requires deep receptive fields and large training datasets. Operator learning approaches, such as Fourier Neural Operators (FNOs) \cite{fno}, address long-range dependencies more explicitly in the frequency domain \cite{choi2024applications}, but inherently introduce additional assumptions, including periodic boundary conditions, homogeneous coefficients, and fixed domain sizes, which limit their applicability in many real-world settings.

Physics-informed ML aims to mitigate data scarcity by embedding physical constraints or priors into the learning process. 
Physics-Informed NNs (PINNs) \cite{dissanayake_neural-network-based_1994,lagaris_artificial_1998,PINN19}, for instance, replace or augment data losses with squared differential equation residuals, enabling learning with limited labeled data. Despite notable progress \cite{cuomo2022scientific, pinn_rao, pinn_sun, grimm_learning_2025, wen2022u, cai_physics-informed_2021}, such methods remain at an early stage of methodological maturity and often struggle in complex, high-dimensional scenarios, for example by converging to trivial solutions \cite{krishnapriyan2021characterizing}.

In our approach, rather than increasing architectural complexity, we build on a simple, well-suited base architecture and exploit the physical structure, i.e., of the underlying advection–diffusion system, explicitly.
Such systems naturally decompose into processes governed by predominantly local interactions, such as diffusion acting at neighborhood scales and inferable from local information, as well as global transport effects, such as long-range advection, that require propagating information across large parts of the domain.
Building on this observation, we propose the \textbf{Local–Global CNN (LGCNN)}, a modular architecture that separates local learning from explicit global transport modeling. LGCNN combines CNNs that are well-suited for locally dominant, diffusion-driven processes with a lightweight numerical surrogate that captures long-range transport effects.
By inserting this inexpensive, physically motivated component at a critical point in the inference pipeline, LGCNN captures long-range dependencies without relying on deep architectures, large receptive fields, or extensive training data. In contrast to physics-informed approaches that softly impose physical constraints through loss terms, such as PINNs,
LGCNN handles global transport at inference time via a dedicated, non-learnable module, leading to improved data efficiency and robustness.

We evaluate LGCNN on the task of predicting city-scale subsurface temperature fields induced by one hundred interacting groundwater heat pumps in the Munich region, Germany~\cite{geokw22}. This problem provides a challenging testbed for data-efficient learning, as the temperature field depends on heterogeneous subsurface parameters, arbitrary source configurations, and long-range advective transport over several kilometers and decades. We show that, even when trained on fewer than five simulated samples, LGCNN generalizes to much larger spatial domains without retraining. Our approach also transfers from synthetic inputs to real subsurface parameter maps derived from borehole measurements.

\begin{figure*}[!tbp]
    \begin{center}
        \includegraphics[trim=40 0 0 0, clip, width=\linewidth]{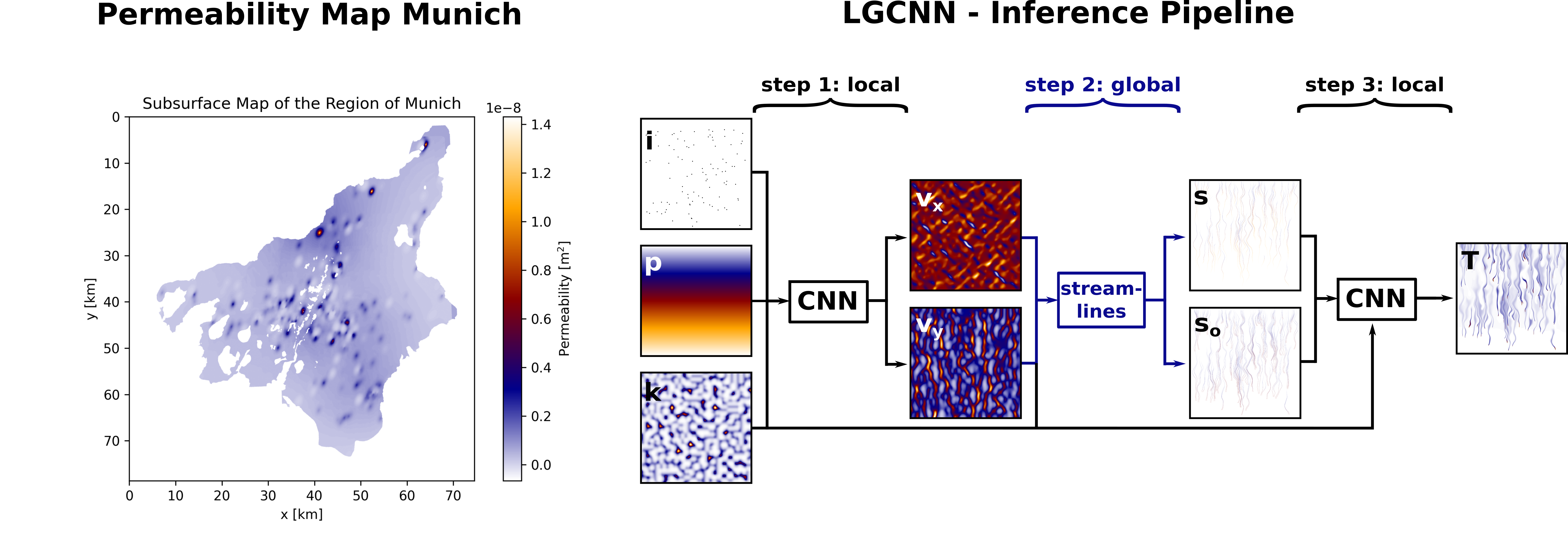}
        \caption{\small Left: Map of $k$ of whole region of Munich. Right: Schematics of our Local-Global CNN-based approach (LGCNN) with 3 physics-inspired steps: CNN ($pki \rightarrow \Vec v$), simplified solver ($i \Vec v \rightarrow \Vec s$), CNN ($pki \Vec v \Vec s \rightarrow T$).}
        \label{fig:perm_map+schematics}
    \end{center}
\end{figure*}
Formally, the learning task is to predict a temperature field $T$ over arbitrarily large domains from heterogeneous permeability fields $k$, hydraulic pressure gradients $\nabla p$, and heat pump locations $i$. Training labels are generated offline using expensive numerical simulations for only a small number of configurations. As illustrated in \Cref{fig:perm_map+schematics}, LGCNN addresses this setting via a three-stage pipeline: (1) a CNN predicts the velocity field $\vec v$ from local relations between $p$, $k$, and $i$; (2) a cheap numerical surrogate computes streamlines $\Vec s$ to model global transport; and (3) a second CNN predicts the temperature field $T$ conditioned on inputs and intermediate representations. This explicit separation of local learning and global transport is key to the scalability of the approach. 

In groundwater heat pump (GWHP) modeling, existing work \cite{davis2023deep, pelzer2024, Stefania} focuses on isolated or pairwise pump interactions, often using UNet-based architectures with or without physics-based regularization. These methods typically require substantial training data and have not been demonstrated to scale to city-wide domains with many interacting sources and heterogeneous aquifers.

\paragraph{Contributions}
We introduce a modular, physics-inspired architecture that captures complex, multiscale dynamics by combining local convolutional feature learning with explicit long-range transport modeling.
To avoid deep architectures and large datasets for modeling nonlocal interactions, LGCNN enforces global coupling through inexpensive numerical transport modules. 
This structural inductive bias enables inherent generalization and scalable inference across domains and problem sizes from very limited data, i.e., less than five simulations.
Compared to classical high-fidelity simulations, our approach achieves an inference-time speedup of approximately 2{,}000 times.

\paragraph{Limitations}\label{sec:limitations}
The current implementation targets a two-dimensional steady-state setting and assumes weak or spatially localized temperature-induced flow perturbations. 
While this work is restricted to this regime, the proposed framework is easily extendable to three-dimensional and transient problems. 
Empirical validation of these extensions is left to future work due to the lack of suitable training data.

\section{Datasets and Metrics}

\paragraph{Datasets}\label{sec:datasets}
Inputs for the NNs consist of a heterogeneous permeability field $k$, an initial hydraulic pressure field $p$, and a one-hot-encoded field of heat pump positions~$i$ of a constant injection rate.
The (interim) labels of velocity~$\Vec v$ and temperature~$T$ fields are simulated with Pflotran \cite{pflotran} until a quasi-steady state is reached after $\approx$27.5~years simulated time \cite{bawue_arbeitshilfe}.

We generate two types of datasets, one is based on synthetic, the other on real permeability fields $k$.
Both cover a $12.8\times12.8$~km$^2$ domain with $2\,560\times2\,560$ cells. The \textbf{baseline dataset} uses Perlin noise~\citep{perlin} to generate random, heterogeneous $k$ fields.
Three simulations (\textit{3dp}) with different fields for $i$ and $k$ are run, generating one datapoint each for training, validation, and testing, plus a 4$\times$ larger simulation to assess scalability.
For training with larger datasets as a reference in \Cref{sec:general_approaches}, we generate an additional dataset of 101~datapoints (\textit{101dp}), split into 73--18--10 for train, val, test data.
Simulation times are $\approx 27$~hours per datapoint and 123~hours for the larger one.

The more \textbf{realistic dataset} builds on $k$ fields cut from maps of borehole measurements in the Munich region, Germany \cite{lfu_gepo}. This dataset consists of four simulations (\textit{4dp}) (three for training and one for validation) and one larger simulation for testing scalability. Due to constraints of availability of subsurface measurements, see \Cref{fig:perm_map+schematics}, the large-scale simulation is on a rectangular domain of only twice the length of the training data. Runtimes range from 38~to 91~hours (average 58~hours), and 134~hours for the scaling-test domain.
Variation within the dataset for training stems from different heat pump placements in $i$ and from the location where $k$ is extract from the Munich region in \ref{fig:perm_map+schematics}.

Information on simulations and hydro-geological parameters are in \Cref{app:params_sim}; hardware specifications in \Cref{app:hardware_specification}.

\paragraph{Performance Metrics}
We evaluate model accuracy separately for each output field (directional velocities $v_x, v_y$, and temperature $T$) using standard ML metrics of Mean Absolute Error (MAE), 
Root Mean Squared Error (RMSE), 
and Huber loss;
the structure-focused metric of Structural Similarity Index Measure (SSIM) \cite{ssim},
and application-driven metrics of Measure of Connectivity (MoC) 
and visual assessment.
MoC expresses how physically reasonable the temperature fields are by measuring the percentage of predicted heat plume cells that are not connected via 4-connectivity\footnote{This means axis-aligned neighbors only, excluding diagonal connections.}
to a heat pump based on a temperature threshold of $0.1$~\degree C, corresponding to measurement precision \cite{kabellichtlot}.
All metrics other than SSIM are applied after re-normalization to the original data ranges to obtain physically meaningful results in [\degree C] or [meters/year].

\section{Application of Established ML Approaches}
\label{sec:general_approaches}
\begin{figure*}[tbp]
    \centering
    \includegraphics[width=\textwidth]{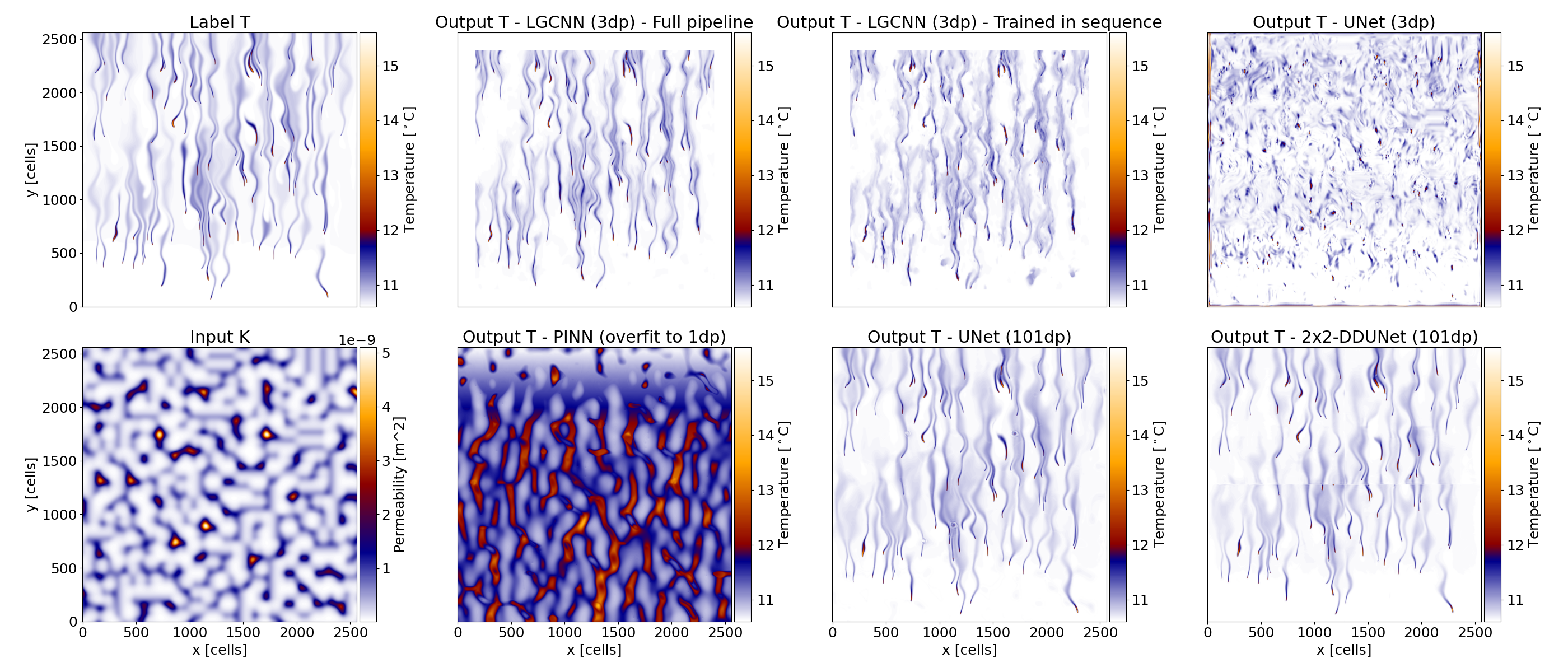}
    \caption{End-to-end: $pki \rightarrow T$ (baseline dataset (\textit{3dp}): test). 1st column: Label $T$ and input $k$. 2nd-4th column: LGCNN$_\textit{3dp}$ (full pipeline and trained in sequence, see \Cref{sec:results}), standard UNet$_\textit{3dp}$, PI-CNN$_\textit{overfit}$, UNet$_\textit{101dp}$, and DDUNet$_\textit{101dp}$ (see \Cref{sec:general_approaches}).}
    \label{fig:dummyK_pki_T}
\end{figure*}

To motivate a structure-specific architecture, we present the performance of six established ML approaches for predicting $T$ from $p, k, i$: a UNet, a domain decomposition-based UNet (DDUNet) \cite{DDUNet}, a dilated UNet, UNet++ \cite{unetpp}, an FNO and a PINN (results in \Cref{tab:results_all,fig:dummyK_pki_T}, parameters in \Cref{app:hyperparam_optimization}).

\paragraph{PINNs and FNOs} 
PINNs, while promising for low-data fluid dynamics tasks \cite {pinn_rao, pinn_sun, cai_physics-informed_2021, pdebench},
struggle in preliminary tests with complex scenarios \cite {krishnapriyan2021characterizing}, such as heterogeneous media and discontinuous source terms.
Physics-informed CNNs (PI-CNNs) have been evaluated in simplified single-plume aquifer settings \cite{hirche23, ymmor24} and consistently failed to outperform standard CNNs, even when the physics-based loss was used only as an auxiliary regularization term alongside a data loss.
The largest errors occur near injection locations, likely due to discontinuous source term fields.
The largest errors occur near injection locations, likely due to discontinuous source term fields.
The largest errors occur near injection locations, likely due to discontinuous source term fields.
In our \textit{3dp} problem, overfitting a PI-CNN to a single datapoint using only energy-based loss terms (cf. \Cref{app:pinn_loss}) derived from the governing equations of the simulations fails to model $T$ but instead predicts a field that resembles $v_y$ (\Cref{fig:dummyK_pki_T}). This behavior is consistent with the model capturing only a preliminary stage of the physical process and indicates a representational limitation of the learning setup regarding transport modeling.

While FNOs are in principle capable of modeling both short- and long-range interactions, we found them impractical for the target problem due to infeasible memory requirements at realistic domain sizes and their sensitivity to non-periodic boundary conditions.
Additionally, they performed poorly in the presence of multiple sources, consistent with~\citet{liu2024neural}.
On our example, we had to reduce the domain size to 480$\times$480 cells to fit it on a graphics processing unit (GPU) (\Cref{app:hardware_specification}) during training and still only managed to predict low frequency aspects, i.e., the background temperature, in our overfitting test to one datapoint.

\paragraph{UNet-based Methods}
Convolutional architectures such as UNet primarily capture local spatial dependencies and can, in principle, generalize from a single large-domain sample when interactions are local. Modeling global effects, however, requires enlarged receptive fields or architectural adaptations, such as dilated convolutions. 
UNet++ fuses multiscale features through nested skip connections, while still relying on convolutional locality.
DDUNet addresses large domains under GPU memory constraints by decomposing the domain into, here $2\times2$, subdomains with learned coarse-scale feature exchange, enabling effective global context at reduced memory cost. 
The hyperparameters of all models are optimized according to \Cref{app:hyperparam_optimization}.

\renewcommand{\arraystretch}{0.95}
\begin{table*}[t]
\caption{\textbf{Metrics of all experiments.} Results for standard models and LGCNN (end-to-end and Steps 1 and 3) on synthetic and real permeability fields $k$, including ablation variants. Models are trained on \textit{3dp}, \textit{101dp}, or \textit{4dp}. Metrics are reported in physical units ([\degree C] for Step 3 and end-to-end, [m/y] for Step 1), except MoC ([\%]) and SSIM (unitless). Velocity experiments are shaded due to differing scales; experiment details are referenced in column Sec. Only contains representative runs, statistical errors are reported in the appendix.}
\centering
\scriptsize
\begin{tabular}{l llllrrrrr|r}
& \textbf{Architecture} & \textbf{Training Data} & \textbf{Applied to} & \textbf{Step} & \textbf{MAE} & \textbf{RMSE} & \textbf{Huber}  & \textbf{SSIM} & \textbf{MoC} & \textbf{Sec.}\\
\midrule
\multirow{6}{*}{\rotatebox{90}{\textbf{Standard}}}
& UNet &  synth. 101dp  & test &  end-to-end & 0.0473 & 0.1001 & 0.0048 &0.8292 & 0.09 & \ref{sec:general_approaches}\\
& UNet++ &  synth. 101dp  & test &  end-to-end & 0.0918 & 0.1435 & 0.0102 & 0.5331 & 4.06 & \ref{sec:general_approaches}\\
& DDUNet &  synth. 101dp  & test &  end-to-end & 0.0548 & 0.1134 & 0.0063 & 0.8120 & 0.59 & \ref{sec:general_approaches}\\
& \textbf{UNet}  &  \textbf{synth. 3dp} & test &  end-to-end & 0.1908 & 0.6190 & 0.0736 & 0.6671 & \textbf{10.85} & \ref{sec:general_approaches}\\
& Dilated UNet  &  synth. 3dp  & test &  end-to-end & 0.1200 & 0.2236 & 0.0237 & 0.5004 & 10.95 & \ref{sec:general_approaches}\\
& PI-CNN (overfit) & synth. 1dp & train & end-to-end & 0.7532 & 0.8593 & 0.3454 & 0.1793 & - & \ref{sec:general_approaches}\\
\hline
\multirow{4}{*}{\rotatebox{90}{\textbf{Ours}}}
& \textbf{LGCNN}  &  \textbf{synth. 3dp} & test &  end-to-end & 0.0916 & 0.1738 & 0.0146 & 0.6841 & \textbf{0.09} & \ref{sec:performance}\\
& UNet  &  synth. 3dp & test &  3 & 0.0417 & 0.0762 & 0.0029 & 0.8540 & 0.07  & \ref{sec:performance}\\
& \hl{UNet}  &  \hl{synth. 3dp} & \hl{test} &  \hl{1 (vx)} & \hl{22.3178} & \hl{31.1860} & \hl{21.8237} & \hl{0.9739} & \hl{-}  & \hl{\ref{sec:performance}}\\
& \hl{UNet}  &  \hl{synth. 3dp} & \hl{test} &  \hl{1 (vy)} & \hl{32.7444} & \hl{45.0703} & \hl{32.2488} & \hl{0.9733} & \hl{-}  & \hl{\ref{sec:performance}}\\
\hline
\multirow{12}{*}{\rotatebox{90}{\textbf{Experiments}}}
& UNet, in sequence  &  synth. 3dp & test & 3 & 0.0919 & 0.1695 & 0.0139 & 0.6714 & 1.30 & \ref{sec:ablation} \\
& UNet, w/ zero-padding  &  synth. 3dp & test & 3 & 0.0487 & 0.0795 & 0.0031 & 0.7784 & 0.57  & \ref{sec:ablation}\\
& UNet, w/o partitioning  &  synth. 3dp & test & 3 & 0.0429 & 0.0793 & 0.0031 & 0.8349 & 0.29  & \ref{sec:ablation}\\
& UNet  &  synth. 101dp  & test & 3 & 0.0416 & 0.0816 & 0.0032 & 0.8016 & 0.47  & \ref{sec:ablation}\\
& DDUNet  &  synth. 101dp  & test & 3 & 0.0376 & 0.0764 & 0.0029 & 0.8843 & 0.26  & \ref{sec:ablation}\\
& UNet, vary inputs & synth. 3dp & test & 3 & cf. \ref{app:exp_inputs} & cf. \ref{app:exp_inputs} & cf. \ref{app:exp_inputs} & cf. \ref{app:exp_inputs} & cf. \ref{app:exp_inputs} & \ref{app:exp_inputs}\\
& \hl{UNet, w/o partitioning}  &  \hl{synth. 3dp} & \hl{test} &  \hl{1 (vx)} & \hl{23.8584} & \hl{33.8297} & \hl{23.3646}  & \hl{0.9765} & \hl{-}  & \hl{\ref{sec:ablation}}\\
& \hl{UNet, w/o partitioning}  &  \hl{synth. 3dp} & \hl{test} &  \hl{1 (vy)} & \hl{35.9249} & \hl{49.9040} & \hl{35.4289} & \hl{0.9721} & \hl{-}  & \hl{\ref{sec:ablation}}\\
& \hl{UNet}  &  \hl{synth. 101dp}  & \hl{test} &  \hl{1 (vx)} & \hl{12.7036} & \hl{29.9379} & \hl{12.2161} & \hl{0.9437} & \hl{-} & \hl{\ref{sec:ablation}}\\
& \hl{UNet}  &  \hl{synth. 101dp}  & \hl{test} &  \hl{1 (vy)} & \hl{324.7751} & \hl{328.4992} & \hl{324.2751} & \hl{0.9572} & \hl{-} & \hl{\ref{sec:ablation}}\\
& \hl{DDUNet}  &  \hl{synth. 101dp}  & \hl{test} &  \hl{1 (vx)} & \hl{13.6736} & \hl{36.9852} & \hl{13.1829} & \hl{0.9468} & \hl{-} & \hl{\ref{sec:ablation}}\\
& \hl{DDUNet}  &  \hl{synth. 101dp}  & \hl{test} &  \hl{1 (vy)} & \hl{322.0676} & \hl{325.6697} & \hl{321.5676} & \hl{0.9741} & \hl{-} & \hl{\ref{sec:ablation}}\\
\hline
\multirow{6}{*}{\rotatebox{90}{\textbf{Scaling}}}
& DDUNet (vanilla) &  synth. 101dp  & scaling &  end-to-end & 0.0235 & 0.0723 & 0.0025& 0.8874 & \textbf{1.18}  & \ref{sec:general_approaches}\\
& UNet (vanilla) &  synth. 101dp  & scaling &  end-to-end & 0.0202 & 0.0578 & 0.0016 & 0.8914 & \textbf{0.54}  & \ref{sec:general_approaches}\\
& \textbf{LGCNN}  &  synth. 3dp & scaling &  end-to-end & 0.0413 & 0.1189 & 0.0065 & 0.6511 & \textbf{0.06} & \ref{sec:performance}\\
& UNet  &  synth. 3dp & scaling &  3 & 0.0168 & 0.0373 & 0.0007 & 0.9405 & 0.08 & \ref{sec:performance}\\
& \hl{UNet}  &  \hl{synth. 3dp} & \hl{scaling} &  \hl{1 (vx)} & \hl{24.9261} & \hl{34.7004} & \hl{24.4314} & \hl{0.9869} & \hl{-} & \hl{\ref{sec:performance}}\\
& \hl{UNet}  &  \hl{synth. 3dp} & \hl{scaling} &  \hl{1 (vy)} & \hl{26.2795} & \hl{38.2599} & \hl{25.7847} & \hl{0.9932} & \hl{-} & \hl{\ref{sec:performance}}\\
\hline
\multirow{8}{*}{\rotatebox{90}{\textbf{Domain transfer}}}
& LGCNN  &  real 4dp  & val &  end-to-end & 0.0841 & 0.1659 & 0.0137 & 0.6511 & 1.65 & \ref{sec:transfer} \\
& UNet  &  real 4dp  & val &  3 & 0.0175 & 0.0319 & 0.0005 & 0.9405 & 3.30 & \ref{sec:transfer} \\
& \hl{UNet}  &  \hl{real 4dp}  & \hl{val} &  \hl{1 (vx)} & \hl{13.3316} & \hl{16.5833} & \hl{12.8383} & \hl{0.9869} & \hl{-} & \hl{\ref{sec:transfer}} \\
& \hl{UNet}  &  \hl{real 4dp}  & \hl{val} &  \hl{1 (vy)} & \hl{9.0224} & \hl{10.5612} & \hl{8.5327} & \hl{0.9932} & \hl{-} & \hl{\ref{sec:transfer}} \\
& LGCNN &  real 4dp  & scaling &  end-to-end & 0.0394 & 0.0665 & 0.0022 & 0.6755 & 3.50 & \ref{sec:transfer} \\
& UNet  &  real 4dp  & scaling &  3 & 0.0189 & 0.0287 & 0.0004 & 0.8942  & 1.77 & \ref{sec:transfer} \\
& \hl{UNet}  &  \hl{real 4dp}  & \hl{scaling} &  \hl{1 (vx)} & \hl{110.0078} & \hl{116.4903} & \hl{109.5079} & \hl{0.9436} & \hl{-} & \hl{\ref{sec:transfer}} \\
& \hl{UNet}  &  \hl{real 4dp}  & \hl{scaling} &  \hl{1 (vy)} & \hl{17.6051} & \hl{24.8298} & \hl{17.1118} & \hl{0.9905} & \hl{-} & \hl{\ref{sec:transfer}} \\
\bottomrule
\end{tabular}
\label{tab:results_all}
\end{table*}

UNet and dilated UNet are trained on \textit{3dp}; UNet, UNet++ and DDUNet are trained on \textit{101dp} directly (\Cref{app:additional_experimental_results}).
All models are evaluated on the \textit{3dp} test data for direct comparability to LGCNN, some on the scaling datapoint. Training on \textit{3dp} leads to short, disconnected plumes (\Cref{fig:dummyK_pki_T}) with high MoC.
When trained on \textit{101dp}, UNet and DDUNet show robust predictions and reasonable scaling, with DDUNet occasionally hallucinating plumes but offering lower memory usage and shorter runtimes (see \cref{tab:training_measurements}). UNet++ fails due to memory constraints in the scaling test.
 
Overall, well-chosen established approaches can perform reasonably (best is UNet$_{101dp}$: MAE = 0.05\degree C, MoC = 0.09) and some can scale to larger domains, but none performs reasonably well in the data-scarce setting we are interested in (best is UNet$_{3dp}$: MAE = 0.19\degree C, MoC = 10.85). Their limits are clear: they are biased toward local patterns, require large datasets to capture long-range transport, and quickly run into compute and memory constraints at realistic scales. In low-data regimes they yield short, disconnected plumes or hallucinated structures, and scalability can break down (e.g., FNO and UNet++ fail under memory constraints).

\section{Methodology}\label{sec:lgcnn}
When generic ML methods fail in data-scarce settings, performance can be improved by adopting a more task-specific approach whose structure reflects the underlying physics \cite{fernandez2024staged, thodi2024fourier, zhang2020physics, yousefzadeh2025determination, bertels2023physics}. 
Accordingly, this section outlines the relevant physical processes that motivate the proposed LGCNN method and discusses key aspects of its components.

\paragraph{Physics of Groundwater Flow with Heat Transport}\label{sec:physics}
Heat transport induced by heat injections of GWHPs in the subsurface is an advection-diffusion process. Since the Péclet number (Pe), i.e., the advection-diffusion ratio, is high in our application (Pe$\gg$1, more in \Cref{app:peclet}), the system is advection-dominated.
In the subsurface, advection is largely governed by the global hydraulic pressure gradient, driving flow from spatially higher to lower regions. Locally, the flow paths of heat plumes are influenced by the spatial distribution of permeability~$k$, cf.~\Cref{fig:dummyK_pki_T}. As a result, small changes in~$k$ can lead to significant differences in flow paths further downstream, demonstrating high input sensitivity.

\paragraph{One-Way Coupled Approach: Local-Global CNN (LGCNN)} \label{sec:oneway}
In a fully coupled simulation, flow field computation and heat transport along streamlines~\citep{streamline} starting at heat pumps (advection) and their widening (diffusion) influence each other. Therefore, convergence requires that two-way coupling conditions are satisfied, whether using a monolithic or a segregated approach.

By working with steady-state simulated $\Vec v$ as interim outputs during training (cf. \Cref{app:assumptions} \textit{Reasoning for Decoupling Assumptions}), we can split the processes into local (diffusion) and global (advection) dominated ones and hence extract three subsequent steps that are coupled in only one direction.
Our simplified physical pipeline consists of three steps: (1) compute a steady-state flow field $\Vec v$ from initial subsurface parameters $p$ and $k$, with $i$ encoding mass influx around GWHPs; (2) transport injected heat along streamlines governed by $\Vec v$ until quasi-steady state \cite{bawue_arbeitshilfe}; and (3) apply plume widening to these heat paths, informed by soil diffusivity and $\Vec v$, to approximate diffusion effects.
The resulting one-way coupled application of LGCNN (see \Cref{fig:perm_map+schematics}) can formally be described as:

\begin{enumerate}[label=\textbf{Step~\arabic*:}, leftmargin=*, labelsep=0.5em]
  \item \textbf{Velocities (local)}
  We employ a CNN to predict the velocity field $\Vec v=(v_x,v_y)$ from $p$, $k$, and $i$:
  \begin{equation}
    \text{CNN}(p,k,i)=\Vec v.
  \end{equation}

  \item \textbf{Streamlines (global)}
  Based on the predicted velocities $\Vec v$, we compute streamlines $\Vec s$ originating from all pumps in $i$ with an initial value problem (IVP) solver:
  \begin{equation}
    \text{IVP}(i,\Vec v)=\Vec s.
  \end{equation}

  \item \textbf{Temperature (local)}
  A second CNN predicts the temperature field $T$: 
  \begin{equation}
    \text{CNN}(k,i,\Vec v,\Vec s)=T.
  \end{equation}
\end{enumerate}

The model outputs a steady-state temperature field $T$ using the same inputs as a simulation, hence serving as its direct surrogate.

\vspace{-6pt}
\paragraph{CNN Models in the Local Steps}
The local Steps 1 and 3 are approximated by a UNet, same as in other groundwater applications \cite{davis2023deep, pelzer2024}, see \Cref{app:unet_block} for details of our architecture. It is beneficial to omit zero-padding to enforce shift invariance and reliance on purely local information \cite{Islam2020How}.
Due to the input sensitivity of the entire problem, 
the CNN in Step 3 is trained on simulated velocities $\Vec v_\text{sim}$ and streamlines based on these fields. Only during inference, all steps are applied sequentially, i.e., Step 3 uses the outputs of Steps 1 and 2, e.g., by including the predicted $\Vec v_\text{pred}$ as inputs.

The training is accelerated by partitioning the data into overlapping patches, which increases the effective number of datapoints while reducing their spatial size.
Combined with the streamlines' embedding and localizing of global flow patterns for the fully convolutional (locally) acting CNN, this training data enrichment allows the model to train effectively on very little data, partially as little as a single simulation run, while generalizing to unseen and even to larger domains. 
Validation and testing are conducted on independent simulation runs, with the full domain being processed as a whole instead of patches during inference.

The architecture, input choices, and training hyperparameters, including patch size and overlap in the datapoint partitioning, were optimized; see \Cref{app:hyperparam_optimization} for details.

\vspace{-6pt}
\paragraph{Streamline Calculation and 2D-Embedding in the Global Step}
Streamlines are calculated by solving the initial value problem (IVP)
\begin{equation}
    \frac{d y}{dt} = \vec v (y), \text{ with } y(t_0) = y_0,
\end{equation}
with a lightweight numerical solver for each heat pump, where $y_0$ represents the location of a heat pump in $i$. We employ
an implicit fifth-order Runge–Kutta method \cite{scipy}
with 10{,}000 one-day time steps, corresponding to the physical timescale of heat plume formation.

The computed streamlines $y$ represent sequences of 2D positions traversed by a particle injected at $y_0$ under the velocity field $\Vec v$. These paths are embedded into 2D fields along the traversed grid cells by assigning values that decay linearly from 1 to 0, 
reflecting the time required to reach each cell and yielding a time-weighted 2D-embedding.

To address the sensitivity of this process with respect to $k$ and $\Vec v$, we calculate the outer streamlines $s_o$. For this, we perturb each heat pump's location $y_0$ by $\pm$10~cells orthogonal to the global flow direction, i.e., $\nabla p$, and compute the corresponding streamlines.
By incorporating $\Vec s = \left (s, s_o \right )$ as two inputs to Step 3, it obtains a spatial, localized representation of flow paths.
In future work, we aim to explore probabilistic perturbations to compute the mean and standard deviation of streamlines, which, while being computationally more expensive, could be efficiently parallelized on GPUs.

\section{Results } \label{sec:results}
We evaluate LGCNN's performance on the baseline dataset \textit{3dp} for isolated Steps~1 and~3, and for the full pipeline (end-to-end), demonstrating the potential and scalability of our approach.
In \Cref{sec:ablation}, we motivate model design choices through variational experiments. Additional metrics and visualizations can be found in~\Cref{app:additional_experimental_results}.
\Cref{sec:transfer} explains adaptations, performance, and scalability on the more realistic dataset.

\subsection{Performance on the Synthetic Dataset}
\label{sec:performance}

\begin{figure*}[tb]
    \centering
    \includegraphics[width=\textwidth]{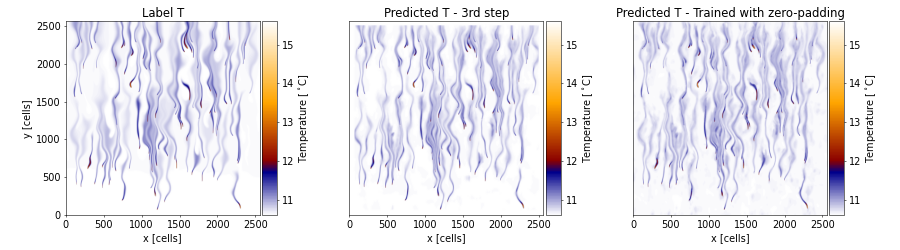}
    \caption{Step 3: $pki\Vec s \Vec v_\text{sim} \rightarrow T$ (baseline dataset (\textit{3dp}): test). 1st column: Label $T$. 2nd-3rd column: LGCNN Step 3, trained without and with zero-padding.}
    \label{fig:results_T}
\end{figure*}

\paragraph{Isolated Steps 1 and 3}
Performance metrics in \Cref{tab:results_all} \textit{Ours} are for Step 3 (trained and tested on simulated groundtruth $\Vec v_\text{sim}$), and for Step 1 highlighted in gray. The combination of quantitative (MAE of (22.3, 32.7)~m/y and 0.04~\degree C, and only 0.07\% unconnected cells) and qualitative measures (visual assessment of \Cref{fig:results_T}) demonstrate a strong, physically reasonable performance, especially compared to a generic UNet trained on the same number of datapoints (\Cref{fig:dummyK_pki_T}).

\paragraph*{Full Pipeline}
Testing the full pipeline, i.e., Steps 2 and 3 applied to $\Vec v_{\text{pred}}$ predicted by Step 1, results in a higher test MAE of 0.09~\degree C compared to an isolated Step~3. This was expected as the isolated Step 3 uses groundtruth, simulated velocity fields $\Vec v_{\text{sim}}$ as inputs.
Visual assessment of representative predictions in~\Cref{fig:dummyK_pki_T} reveals physically plausible heat plumes in terms of shape, extent, and heat magnitude, also measurable by a comparably low MoC of 0.09~\%.
Deviations in streamlines arise from smaller errors in the velocity predictions $\Vec v_{\text{pred}}$ from Step 1, highlighting the input sensitivity of the physical problem, especially near bifurcations, where small perturbations in $\Vec v$ can lead to an alternative path, e.g., around a clay lens in $k$. Compared to training end-to-end on \textit{101dp} (see \Cref{tab:results_all} \textit{Standard}), our model achieves a similar MoC and at least half the accuracy while training on only 1 versus 73 training datapoints, which strongly reduces computation time (see~\Cref{tab:training_measurements} for details) and data requirements.

\begin{figure*}[tb]
    \centering
    \hspace*{-15pt}
    \includegraphics[trim=25 15 15 15, clip, width=\textwidth]{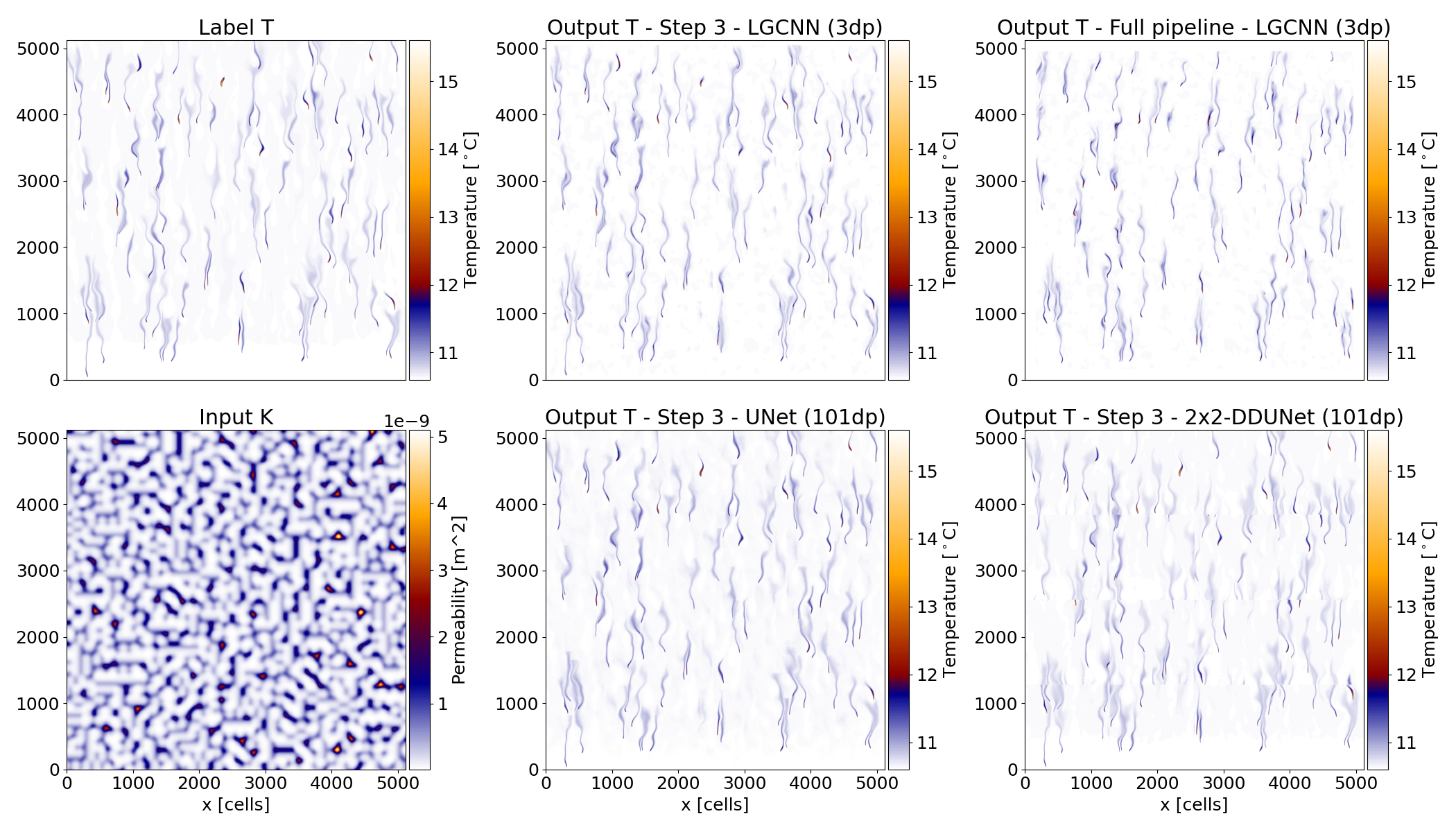}
    \caption[]
    {Predictions of $T$ (baseline dataset (\textit{3dp}): scaling). 1st column: Label $T$ and input $k$. 2nd-3rd column: LGCNN - isolated Step 3 and full pipeline, and prediction for Step 3 based on data-rich training (UNet$_\text{\textit{101dp}}$ and $2\times 2$-DDUNet$_\textit{101dp}$).}
    \label{fig:scalability}
\end{figure*}

\paragraph{Scaling}
It is expected that our model can scale to larger domains, as we train on small patches of the spatial domain and achieve high test accuracies on the full domain during validation and testing; see \Cref{sec:ablation} \textit{Partitioning Training Data} for more details. To further demonstrate the scaling, we test our model on a domain of 4$\times$ the size of the training domain with the same number of heat pumps, 
cf. \Cref{fig:scalability}.
LGCNN obtains a low MAE of 0.02~\degree C for Step 3, and 0.04~\degree C for the full pipeline and a very low MoC of 0.06\%, see \Cref{tab:results_all}.
A visual assessment of \Cref{fig:scalability} yields similar qualitative, physically reasonable behavior as the previous results in \Cref{fig:dummyK_pki_T}, demonstrating the scalability of LGCNN.

\subsection{Experiments on the Synthetic Dataset}\label{sec:ablation}
\paragraph{Training in Sequence}
When training Step~3, we consider two alternatives for the velocity input: simulated velocities $\vec v_{\text{sim}}$ and velocities predicted by Step~1 $\vec v_{\text{pred}}$. Training the model sequentially using $\vec v_{\text{pred}}$ results in substantially higher MoC values compared to training with $\vec v_{\text{sim}}$, as shown in \Cref{tab:results_all} (\textit{Experiments} vs.\ \textit{Ours}).
As illustrated in \Cref{fig:dummyK_pki_T}, using $\vec v_{\text{pred}}$ during training leads to physically implausible temperature fields characterized by noisy artifacts and fragmented heat plumes. We attribute this behavior to local misalignments between the predicted streamline inputs~$\vec s$ and the corresponding temperature label $T$, which strongly complicates localized learning.
In contrast, training with $\vec v_{\text{sim}}$ produces streamline fields that are physically consistent with the temperature distribution which provides informative gradients and enables stable and effective training.

\paragraph{Enabling Zero-Padding}
Using no padding in all convolutional layers produces smaller but cleaner output fields, as shown in \Cref{fig:results_T}. This leads to a slight improvement in standard ML metrics and a substantial improvement in the MoC score (\Cref{tab:results_all}, \textit{Ours} vs. \textit{Experiments}), indicating fewer disconnected heat plumes. 

\paragraph{Partitioning Training Data}
Hyperparameter search shows the best performance for partitioning the domain during training into 20\,736 overlapping patches of 256$\times$256 cells for Step 1 and 82\,944 patches for Step 3. Although this increases per-epoch compute time compared to full-domain training, it significantly reduces the number of epochs needed to reach comparable test performance (cf.~\Cref{tab:results_all} \textit{Experiments} vs. \textit{Ours, Step 1}), resulting in an overall training time reduction of 67--90\% (cf.~\Cref{tab:times}) while maintaining comparable accuracy across all metrics. 

\paragraph{Replacing Isolated Steps with (2$\times$2-DD)UNet$_\textit{101dp}$}
We replace each step with an optimized UNet or DDUNet trained on \textit{101dp} (cf.~\Cref{app:hyperparam_optimization}). 
Step~2 fails for both models. 
Results for Steps~1 and~3 are summarized in~\Cref{tab:results_all} \textit{Experiments}. In Step~1, both models struggle to predict $v_y$ due to outliers in training data. In Step~3, performance is comparable to \textit{3dp} in standard metrics but yields a substantially higher MoC, with scaling results in \Cref{fig:scalability} showing minimal benefit from increased training data.
 
\paragraph{Modifying Streamline Inputs of Step~3}
We evaluate the influence of Step 3's inputs, focusing on the effect of modifying or omitting individual components of the streamlines $\Vec s = \left (s, s_o \right )$. When both the central streamlines $s$ and offset streamlines $s_o$ are excluded, predicted plumes fail to follow flow paths beyond the CNN's receptive field. Omitting only $s_o$ results in overly narrow plumes, whereas excluding only $s$ produces overly smeared temperature fields.
Including both but removing the fading over time along the streamlines leads to diffuse background temperatures and overly precise and too long plumes. Overall, the Huber validation loss increases by 32-132\%, depending on the experiment. For extended and visual results, we refer to \Cref{app:exp_inputs}.

\subsection{Domain Transfer to Real Permeability Fields}\label{sec:transfer}
We evaluate our method on real permeability fields $k$ (\Cref{sec:datasets}), for which only four datapoints are available. 
Results for Step~3 and the full pipeline on the validation and scaling datapoints are shown in \Cref{tab:results_all} \textit{Domain transfer}; additional metrics are provided in \Cref{app:additional_experimental_results}.

\begin{figure}[!htbp]
    \centering
    \includegraphics[trim=25 15 15 15, clip, width=\columnwidth]{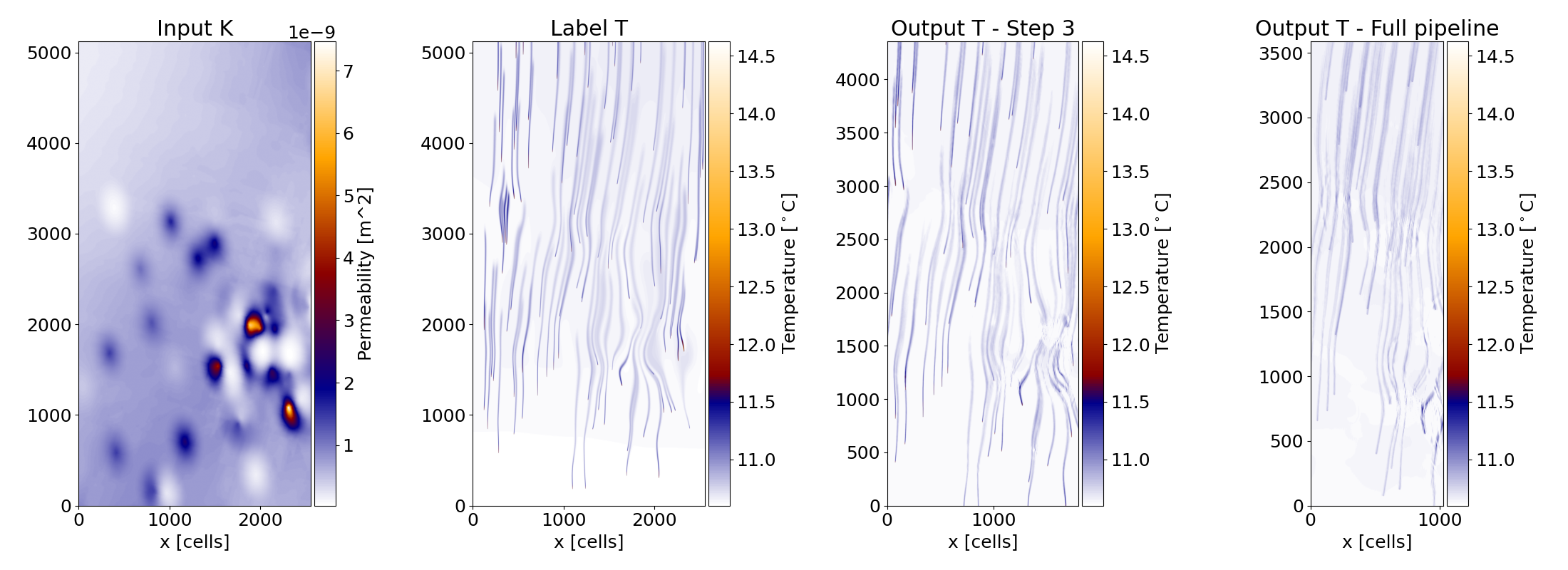}
    \caption[]
    {Predictions of $T$ (realistic dataset (\textit{4dp}): scaling). Input $k$, label $T$ LGCNN isolated Step 3, and full pipeline.}
    \label{fig:scalability_real}
\end{figure}

\paragraph{Methodological Adaptations}
Slight adjustments to chunk size, overlap, data split, and architecture yield better performance compared to those used on the baseline dataset, see \Cref{app:hyperparam_optimization} for details.
Visual inspection of input $k$ in~\Cref{fig:scalability_real} reveals fewer but larger-scale features. Therefore,
training can benefit from a larger spatial context, which requires more training data. We increase the number of training samples by using three out of the four available datapoints for training and one for validation.
For the streamlines computation, we switch to an explicit fourth-order Runge--Kutta scheme, which proved stable and sufficient for training and inference due to the lower frequencies in $k$.

\paragraph{Performance and Scaling Tests}
In Step~3, we reach an even lower maximum absolute error $L_{\infty}$ of 0.82~\degree C compared to 2.90~\degree C of our initial model on the baseline dataset (\Cref{sec:results}). Other metrics confirm this except for SSIM and MoC; cf. \Cref{tab:results_all} \textit{Domain transfer}. In the scaling test, most errors behave as before, as expected.
For the full pipeline, errors are slightly higher on the validation and scaling data compared to an isolated Step~3, consistent with our observations on the baseline dataset.
Physical consistency, which is our primary objective, remains strong: Shape, magnitude, and connectivity of the predicted heat plumes are well preserved, even in the scaling test (\Cref{fig:scalability_real}).
Local deviations are primarily due to differences between $\Vec v_\text{sim}$ and $\Vec v_\text{pred}$, as evident from the discrepancy between Step~3 and full pipeline outputs.

Based on their poor generalization in data-scarce regimes observed on the baseline dataset, we omit generic ML architectures from the experiments on realistic $k$ fields.

\section{Conclusions and Future Work}
We introduce the \textbf{Local-Global Convolutional Neural Network (LGCNN)} to address complex, data-scarce learning problems governed by both local and non-local physical interactions.
Trained on only a few simulations, LGCNN accurately predicts subsurface temperature fields on much larger two-dimensional heterogeneous domains, while providing an inference-time speed-up of roughly 2,000× compared to classical high-fidelity simulations on the training data domain. 
By combining translation-invariant CNNs for local processes with a lightweight numerical surrogate for global transport, our method achieves data-efficient, scalable, and physically consistent predictions, outperforming conventional ML models such as UNet variants that struggle to capture long-range dependencies in data-scarce regimes. 
Although demonstrated for groundwater heat transport, the concept of local-global decomposition can be generalized to a broad class of advection-diffusion systems. 
Future work will extend the approach to three-dimensional and transient scenarios, investigate sensitivity to input uncertainty, and further accelerate streamline computation.

\onecolumn
\newpage

\section*{Acknowledgements}
We thank the Deutsche Forschungsgemeinschaft (DFG, German Research Foundation) for supporting this work by funding - EXC2075 – 390740016 under Germany's Excellence Strategy. We acknowledge the support by the Stuttgart Center for Simulation Science (SimTech). We also acknowledge the use of computational resources of the DelftBlue supercomputer, provided by Delft High Performance Computing Centre~\cite{DHPC2024}.
We thank Kai Zosseder and Fabian Böttcher at the Technical University of Munich for their application-related input.

\section*{Reproducibility}\label{sec:data_avail}\label{sec:reproducibility}
To ensure reproducibility, our code to train and evaluate models, the trained and evaluated models themselves, and the datasets designed for training and testing are included in the supplementary material and will be published after the review process.
Our raw data and pretrained models are publicly available on SurfDrive, see below in \textit{Supplementary Material}.

Our code for data generation (\url{https://github.com/JuliaPelzer/Dataset-generation-with-Pflotran}) and model training (including dataset preparation and evaluation routines) (for LGCNN and data preparation: \url{https://github.com/JuliaPelzer/Heat-Plume-Prediction/tree/AllIn1/LGCNN/release25}, for DDUNet: \url{https://github.com/corne00/DDUNetForHeatplumePrediction}) is on GitHub. Follow the instructions in the respective readme files to prepare the data and train or infer a model.


\section*{Supplementary Material}\label{sec:suppl}
The supplementary material contains the raw datasets, the most important trained models and the code basis for preparing the raw data to train on, separate training routines and evaluation protocols for LGCNN (on real or synthetic/random permeability fields), UNet$_\textit{3dp}$, experiments with \textit{3dp}; and on the other hand everything with \textit{101dp}: DDUNet$_\textit{101dp}$, $UNet_\textit{101dp}$, experiments with \textit{101dp}.

Raw datasets:
\begin{itemize}
    \item Dataset of random permeability with 3+1 datapoints (\textit{3dp} + 1)
    \item Dataset of random permeability with 101 datapoints (\textit{101dp})
    \item Dataset of real permeability with 4+1 datapoints (\textit{4dp} + 1)
    \item Prepared dataset for training the FNO, because the other datapoints are too large
\end{itemize}

Trained models (including hyperparameters):
\begin{itemize}
    \item vanilla approaches trained on random permeability fields, \textit{3dp}: UNet$_\textit{3dp}$, dilated UNet$_\textit{3dp}$
    \item vanilla approaches trained on random permeability fields, \textit{101dp}: DDUNet$_\textit{101dp}$, UNet$_\textit{101dp}$, UNet++$_\textit{101dp}$
    \item LGCNN on random permeability fields, \textit{3dp}
    \item LGCNN experiment: replace isolated steps 1 and 3 with DDUNet$_\textit{101dp}$, UNet$_\textit{101dp}$
    \item LGCNN on real permeability fields, \textit{4dp}
    \item vanilla models first trying to overfit them on 1 data point:  PI-CNN, FNO
\end{itemize}

Code (including training and evaluation routines):
\begin{itemize}
    \item Repository of 101dp-vanilla approaches (DDUNet$_\textit{101dp}$, UNet$_\textit{101dp}$, also experiment "replace isolated steps")
    \item Repository LGCNN + UNet$_\textit{3dp}$ (including preparation script for datasets to prepare datasets, for all models and approaches)
    \item Repository UNet++$_\textit{101dp}$
\end{itemize}
All supplementary material can be accessed via SURFDrive\\ under this link: \href{https://surfdrive.surf.nl/files/index.php/s/f3Oqg3ufir9T9LL}{https://surfdrive.surf.nl/files/index.php/s/f3Oqg3ufir9T9LL}.


\bibliography{library.bib}
\bibliographystyle{icml2026}

\newpage
\appendix

\newpage
\section{Appendix}\label{sec:appendix}

\subsection{Simulation Setup and Transport Regime}\label{app:simulation_setup}
In~\Cref{sec:datasets}, we describe how the two datasets are generated using the subsurface simulation software Pflotran~\citep{pflotran}, which solves the coupled mass and energy conservation equations. Here, we provide additional technical details and modeling assumptions relevant for reproducibility. Furthermore, we show that for the selected parameters, the heat transport in our system is dominated by advection by a theoretical analysis.

\paragraph{Hydro-geological, Operational and Simulation Parameters}\label{app:params_sim}
Our data was generated on 12.8km$\times$12.8km$\times$5m with a cell size of 5m$\times$5m$\times$5m for the baseline simulations. To test scalability, we also simulate a domain that is twice as large in both $x$- and $y$-dimension for the baseline dataset, but only twice as large in $y$-dimension for the more realistic dataset due to dimension restrictions of the available input data. 

The \textit{baseline dataset} uses a constant hydraulic pressure gradient $\nabla p$ of 0.003~\citep{pelzer2024,geologica}; permeability field $k$ is generated using Perlin noise~\citep{perlin} within (1.02E-11, 5.10E-9)~m$^2$, and $100$ randomly placed heat pumps, which are all operating with a 5~\degree C injection temperature difference compared to the surrounding groundwater and an injection rate of 0.00024~m$^3$/s. All values are based on realistic parameter ranges in the region of Munich\citep{geologica,geokw22}.

For the \textit{realistic dataset}, instead of random permeability fields $k$, we use permeability field data that is derived from borehole measurements in the Munich region\citep{lfu_gepo}. Furthermore, based on subsurface measurements, we set $\nabla p = 0.0025$ for these simulations. 
All other simulation parameters remain identical to the baseline. Other subsurface and operational parameters are taken directly from~\citet{pelzer2024}, which also includes additional information about solver setup and boundaries.
For mathematical details of the governing equations, we refer the reader to~\citet{pflotran_theory, anderson2007introducing, delleur2016elementary}.

\paragraph{Notation Simplification}
For simplifications, we refer to the (hydraulic) pressure field $p$ throughout the paper. In our paper, this field is the initial pressure field defined through the measured hydraulic head and is not the true pressure field at quasi-steady state with spatial details resulting from permeability $k$ variations in the domain and mass injections at the heat pump positions. The true pressure field is only available after simulation (which we are avoiding for our approach) and strongly interacts with the velocity field.

\paragraph{Péclet Number}\label{app:peclet}
To quantify whether heat transport in our system is dominated by advection or diffusion, we compute the dimensionless Péclet number~\citep{rapp2017peclet}, which is defined as
\[
\text{Pe} = \frac{L \cdot v}{\alpha},
\]
with the characteristic length \(L\), the local velocity \(v\), and the thermal diffusivity \(\alpha\), defined as
\[
\alpha = \frac{\kappa}{\rho c_p},
\]
with $\kappa$ the thermal conductivity, \(\rho\) the density, and \(c_p\) the specific heat capacity. We take the parameters of the solid phase of our simulation inputs of
\begin{itemize}
    \item Thermal conductivity: \(\kappa = 0.65\, \mathrm{W/(m \cdot K)}\)
    \item Density: \(\rho = 2800\, \mathrm{kg/m^3}\)
    \item Specific heat capacity: \(c_p = 2000\, \mathrm{J/(kg \cdot K)}\)
\end{itemize}

This yields a thermal diffusivity of:
\[
\alpha = \frac{0.65}{2800 \cdot 2000} \approx 1.16 \times 10^{-7}\, \mathrm{m^2/s}
\]

The velocity values are derived by simulations, taken in the direction of flow (longitudinal) from the realistic $k$-dataset:
\begin{itemize}
    \item Maximum: \(1200\, \mathrm{m/year} \approx 3.8 \times 10^{-5}\, \mathrm{m/s}\)
    \item Minimum: \(44\, \mathrm{m/year} \approx 1.4 \times 10^{-6}\, \mathrm{m/s}\)
    \item Mean: \(330\, \mathrm{m/year} \approx 1.04 \times 10^{-5}\, \mathrm{m/s}\)
\end{itemize}

With a characteristic length determined by the heat plume length of \(2000\text{--}6000\, \mathrm{m}\) and mean x-velocity, we get:
\begin{equation*}
    Pe = \frac{L \cdot v}{\alpha} \approx L \cdot 10^2\ \approx 10^5
\end{equation*}

The interpretation of the Peclet number is given by the following:
\begin{itemize}
    \item $Pe \ll 1$: Diffusion dominates
    \item $Pe \approx 1$: Diffusion and advection similarly dominate
    \item $Pe \gg 1$: Advection dominates
\end{itemize}
Therefore, \(\text{Pe} \approx 10^5\) indicates that in the simulations, the heat transport is advection-dominated at larger scales.

\paragraph{Reasoning for Decoupling Assumptions}\label{app:assumptions}
In LGCNN, the goal is to split properties into local and global. To build separate datasets for the isolated steps, we require reasonable inputs and labels.
Hence, the from data or simulation generated physical properties are analyzed as potential interim inputs / labels.
From simulation, we get $\Vec v$ (mostly local) and $T$ (heat is transported in an advective and diffusive manner $\rightarrow$ local and global). To split the temperature field into local and global, we handcraft a new property streamlines $\Vec s$ that describe the central evolution of a heat plume driven only by advection, starting at each heat pump location.
An important assumption underlying our approach is that the coupling between velocity and temperature can predominantly be treated as one-way in our setting. The velocity field strongly influences the temperature distribution but the temperature also influences the velocity. Since this effect mostly occurs in 3D domains with very high temperature gradients and we model 2D and very small temperature gradients, our assumption holds.

The general background velocity field influences the streamlines directly but the streamlines or rather the injection points and their mass influx also disturbs the local velocity field which in turn can influence other streamlines passing by from further upstream. By including the start locations of the streamlines, i.e., the locations of the heat pumps, as inputs to $\Vec v$ and taking the steady-state flow field with running heat pumps from the simulation as a label, the influence of $\Vec s$ is reduced and again decoupled as an additional input $i$ with the influence only local and already included in the labels. Hence, a one-way coupling from positions of heat pumps $i \rightarrow \Vec v \rightarrow \Vec s$ is possible.

\subsection{NNs, Hyperparameters and Training Details}\label{app:params_nn}
This section provides implementation and training details for all NN models used in this work. We begin with a description of the baseline UNet architecture, which also forms the foundation of both LGCNN and DDU-Net. Then, we outline the hyperparameter optimization process and present the final configurations per model and dataset. All models are trained using PyTorch~\citep{paszke2017automatic}.

\subsubsection{UNet Architecture}\label{app:unet_block}
The UNet architecture used in this work is based on the original design presented by \citet{ronneberger2015u}, with several modifications to tailor it to the specific needs of our task of predicting temperature fields. In this section, we introduce the key concepts that define the architecture and explain how they contribute to the model's design. These concepts will be employed in the hyperparameter search to identify the most suitable configuration, taking into account memory and data limitations.

The UNet architecture essentially consists of a series of UNet blocks. Each block consists of the layers of "Convolution - Activation - Convolution - Norm - Activation - Convolution - Activation" with convolutional layers, a batch normalization layer, and activation functions. After each block, either an upsampling or a downsampling operation is applied via "Max Pooling 2D" or "Transposed Convolution 2D" with stride 2. The \textit{depth} of the UNet refers to the number of UNet blocks in both the encoder and decoder. The \textit{number of initial features} refers to the number of feature maps generated by the first downsampling block. Each downsampling block produces twice as many output feature maps as input feature maps, while each upsampling block reduces the number of feature maps by half. The \textit{number of convolutions per block} denotes how many convolutional layers are applied within each block, while \textit{kernel size} specifies the size of the kernels used in the convolutional operations. Additionally, we explore different activation functions (ReLU, tanh, sigmoid, and LeakyReLU) and various normalization strategies (batch normalization, group normalization, and no normalization).
The UNet block exists in 2 variants, depending on the hyperparameter \textit{repeat inner}: If \textit{repeat inner} = False, the block looks like this "Convolution - Norm - Activation", if it is True, the block looks as described above.

The training process also involves several hyperparameters. The learning rate controls how quickly the model adjusts its parameters during training. The weight decay parameter helps prevent overfitting by penalizing large weights in the model. Furthermore, the Adam optimizer~\citep{kingma2014adam} is employed as \textit{optimizer}. For the realistic permeability field, we additionally introduce the optimizable hyperparameter \textit{optimizer switch}, which, when enabled, switches the optimizer to limited-memory BFGS (LBFGS) after 90 epochs.

During inference, each datapoint is processed as a whole, but during training, they are loaded in smaller patches cut out from the datapoint. Optimized hyperparameters include their overlap, i.e., inverse of \textit{skip per direction}, and the patch size, called \textit{box length}.
The data split is untouched by this, i.e., separate simulation runs for train / val / test. This is important to not overlap test patches with training ones.

\subsubsection{Energy-Based Loss for PINNs}\label{app:pinn_loss}
For our PINN, we overfitted a UNet with an energy loss $\mathcal{L}_\text{energy}$ to a single datapoint from \textit{3dp}.

The physics-informed loss for predicting the temperature field $T$ considers energy conservation, including advective and conductive terms, and heat pump contributions. Let $\vec{v} = (v_x, v_y)$ be the velocity field and $p$ the pressure. Denote the predicted temperature as $\hat{T}$. The energy loss is defined as

\begin{equation}
\mathcal{L}_\text{energy} = \text{MSE}\Big( E - E^\text{HP}_\text{conv}, \, 0 \Big),
\end{equation}
where $E$ is the local energy residual and $E^\text{HP}$ the energy inflow from heat pumps.

\paragraph{Energy residual.}  
The energy residual accounts for advection and thermal diffusion:

\begin{align}
\rho_m &= \text{molar density}(\hat{T}, p), \\
h &= \text{enthalpy}(\hat{T}, p), \\
E_u &= \frac{\partial}{\partial x} \Big( \rho_m \, h \, v_x - \frac{\kappa}{\Delta x} \frac{\partial \hat{T}}{\partial x} \Big), \\
E_v &= \frac{\partial}{\partial y} \Big( \rho_m \, h \, v_y - \frac{\kappa}{\Delta x} \frac{\partial \hat{T}}{\partial y} \Big), \\
E &= E_u + E_v,
\end{align}
where $\kappa$ is the thermal conductivity and $\Delta x$ is the uniform grid spacing.

\paragraph{Heat pump contribution.}  
Heat pumps add localized energy sources based on a one-hot-encoding of the positions of heat pumps (ID) to get a spatially distributed contribution:

\begin{align}
\rho &= \text{density}(\hat{T}, p), \\
E^\text{HP} &= c_p \, \rho \, \Delta T \, Q \, \frac{\text{ID}}{\Delta x^3}, \\
E^\text{HP}_\text{conv} &= K * E^\text{HP},
\end{align}
where $c_p$ is the specific heat capacity of water, $\Delta T$ is the temperature difference at the injection well, $Q$ is the volumetric flow rate, ID is the material indicator (1 if heat pump present, 0 otherwise), and $K$ is a fixed, handcrafted $3\times3$-convolutional kernel $K=\begin{bmatrix}-1&0&1\\0&0&0\\1&0&-1\end{bmatrix}$ with zero-padding used to smooth the inflow contribution. It distributes the inflow over neighboring cells while preserving directional structure and its shape matches our observations.

\subsubsection{Hyperparameters and Hyperparameter Optimization}\label{app:hyperparam_optimization}
We optimize the hyperparameters of our architecture, training process and data loading in several rounds with an automated tree-based search using Optuna~\citep{optuna_2019} and additional manual tweaking. Optuna performs optimization using a tree-structured Parzen estimator algorithm. An overlook of all considered hyperparameters, their ranges and our final choice, as well as the hyperparameters fixed during optimization is provided per used model architecture and dataset, e.g., purely data-driven approaches, LGCNN on individual steps or the full pipeline, on datasets of random $k$ versus realistic $k$.

\paragraph{Vanilla UNet}\label{app:hparams_unet}
The vanilla UNet$_{\textit{3dp}}$ was trained using the following hyperparameters: a batch size of 20, kernel size of 5, and a network depth of 3. The number of initial features was set to 32, with a stride and dilation of 1. We used the ReLU activation function in combination with batch normalization. The inner block was not repeated (\texttt{repeat\_inner = False}). 
No cutouts are applied during training (\texttt{bool\_cutouts = False}), i.e., the model is trained on the whole datapoint at once.

The inputs to the model are $pki$ (pressure field, permeability field, and location of heat pumps). For training, the MAE loss function was used, and optimized with the Adam optimizer. The learning rate is set to $1 \times 10^{-5}$.
The model is trained for 10,000~epochs.

\paragraph{Hyperparameters for UNet variants on 101dp}\label{app:vanilla_ddunet}
The values considered during hyperparameter optimization for the UNet and DDUNet, trained on 73 datapoints and performed using Optuna~\citep{optuna_2019}, along with the best settings found, are listed in~\Cref{tab:params_ddunet}. Certain hyperparameters were fixed: we set the number of epochs to 750, with an early stopping criterion based on validation loss and a patience of 80 epochs. Additionally, we note that some hyperparameter combinations (e.g., 32 initial features, depth 6, and 3 convolutions per layer with a kernel size of 7) caused memory issues, leading to their exclusion from the hyperparameter search.

\begin{table}[!htbp]
\caption{Overview of used hyperparameters for the UNet$_{101dp}$ and ${2\times2}$ DDUNet$_{101dp}$ their search ranges (if applicable), and best values across training stages. Note that the number of communicated feature maps in the vanilla UNet is simply an extra convolution layer in the coarsest part of the UNet (without communication).}
\begin{center}
\resizebox{\textwidth}{!}{
\begin{tabular}{cc|cc|cc|cc}
    \hline
    \textbf{Hyperparameter} & \textbf{Range} & \multicolumn{2}{c}{\textbf{$pki \rightarrow v_x v_y$} (Step 1)} & \multicolumn{2}{c}{\textbf{$i v_x v_y s s_o k\rightarrow T$} (Step 3)} & \multicolumn{2}{c}{\textbf{$pki \rightarrow T$} (Full)} \\
    \hline
    \hline
    & & UNet$_{101dp}$ & ${2\times2}$DDUNet$_{101dp}$ &UNet$_{101dp}$ & ${2\times2}$DDUNet$_{101dp}$ &UNet$_{101dp}$ & ${2\times2}$DDUNet$_{101dp}$\\
    \hline

    \multicolumn{8}{c}{\cellcolor{gray!20}\textit{Dataset}} \\
    Batch size (train)             & 4, 6, 8             & 4      & 6      & 6      & 4      & 6      & 6      \\
    Include pressure field         & True, False         & False  & False  & -  & -  & False      & False      \\
    \hline

    \multicolumn{8}{c}{\cellcolor{gray!20}\textit{Encoder-decoder properties}} \\
    Depth                          & 4, 5, 6             & 6      & 5      & 6      & 6      & 5      & 5      \\
    No. initial features           & 8, 16, 32           & 8      & 16     & 8      & 8      & 8      & 16     \\
    No. convs. per block     & 1, 2, 3             & 1      & 1      & 3      & 3      & 3      & 3      \\
    Kernel size                    & 3, 5, 7             & 7      & 7      & 3      & 5      & 7      & 7      \\
    \hline

    \multicolumn{8}{c}{\cellcolor{gray!20}\textit{Communication Network}} \\
    No. comm. feature maps        & 64, 128, 256        & 64     & 128    & 256    & 64    & 256     & 256    \\
    \hline

    \multicolumn{8}{c}{\cellcolor{gray!20}\textit{Training}} \\
    Learning rate                  & [1e-5, 1e-3]        & 0.00024 & 0.00100 & 0.00017 & 0.00030 & 0.00024 & 0.00024 \\
    Weight decay                   & 0.0, 0.001          & 0.0     & 0.0    & 0.0    & 0.0    & 0.0    & 0.0    \\
    Train loss                     & MSE, L1             & MSE     & MSE    & L1     & MSE    & MSE    & MSE    \\
    \hline
\end{tabular}
}
\label{tab:params_ddunet}
\end{center}
\end{table}

After the hyperparameter search, the values corresponding to the best-performing model (based on Huber loss for the validation dataset) were selected. With these values fixed, five models were trained using different randomly sampled initializations to evaluate sensitivity to random initialization, for these values see~\Cref{tab:model_metrics_pm,tab:modelT_metrics_datadriven_pm}.

\paragraph{Hyperparameters for UNet+++ and dilated UNet}\label{app:vanilla_dilated_pp}
Hyperparameters of our UNet++ and dilated UNet trainings are listed in \Cref{tab:params_dilated_unetpp}.
As far as possible, we stuck to the previously excessively optimized, well-working architecture: For the dilated CNN, we based them on the vanilla UNet \textit{3dp} and only varied the dilation in [2, 4, 8, 10, 16, 32, 100] with 10 producing the lowest validation error and hence listed here.
UNet++ is based on the set of hyperparameters from the UNet++-paper \cite{unetpp} with parameters that were adapted to fit our scenario listed below, based on the hyperparameter search of UNet$_101dp$. For example, the weight decay of the Adam optimizer is set to 0.0 and the training loss to MSE.
The number of epochs for UNet++ is so small due to limited training time on the server. 

\begin{table}[!htbp]
\caption{Baseline hyperparameters for end-to-end temperature prediction.}
\begin{center}
\begin{tabular}{cc|c}
\hline
\textbf{Parameter} & \textbf{Dilated CNN (3dp)} & \textbf{UNet++ (101dp)} \\
    \hline
    \hline
\multicolumn{3}{c}{\cellcolor{gray!20}{\textit{Parameters of the Dataset}}}\\
dataset size & 3 & 101 \\
inputs & $pik$ & $ik$ \\
cutouts / patches & full domain & full domain \\
\hline
\multicolumn{3}{c}{\cellcolor{gray!20}{\textit{Parameters of Training}}}\\
batch size & 20 & 2 \\
training loss & MAE & MSE \\
initial learning rate & $10^{-4}$ (manual decay) & $2\times10^{-4}$ \\
optimizer & Adam & Adam \\
epochs & 1000 & 190\\
\hline
\multicolumn{3}{c}{\cellcolor{gray!20}{\textit{Parameters of the Network}}}\\
architecture & Dilated CNN (zero-padded) & UNet++ \\
kernel size & 5 & \textit{default} \\
depth & 3 & \textit{default} \\
initial features & 32 &  \textit{default}\\
stride & 1 & \textit{default} \\
dilation & 10 & \textit{default} \\
activation function & ReLU & \textit{default} \\
normalization & BatchNorm & \textit{default} \\
repeat inner blocks & False & - \\
\end{tabular}
\label{tab:params_dilated_unetpp}
\end{center}
\end{table}

\paragraph{Hyperparameters for PINN and FNO}\label{app:vanilla_fnopinn}
Hyperparameters of our PI-CNN and FNO trainings are listed in \Cref{tab:params_fno_picnn}.
For PI-CNN, we used the optimized, well-working architecture from before; for FNO a transferred set of hyperparameters with an additional hyperparameter search on layers, modes, inputs and optimizer.

\begin{table}[!htbp]
\caption{Baseline hyperparameters for overfitting experiments on a single datapoint (dp id 0).}
\begin{center}
\begin{tabular}{ccc}
\hline
\textbf{Parameter}  & \textbf{FNO (overfit test)} & \textbf{PINN (overfit test)} \\
\hline
\multicolumn{3}{c}{\cellcolor{gray!20}{\textit{Parameters of the Dataset}}}\\
dataset size & 1 & 1 \\
inputs &  $pik$ & $pikv_xv_y$ \\
box length &  480 & 256 \\
\multicolumn{3}{c}{ }\\
\hline
\multicolumn{3}{c}{\cellcolor{gray!20}{\textit{Parameters of Training}}}\\
batch size &  8 & 20 \\
training loss & MSE & MSE (physics-only) \\
initial learning rate & $10^{-3}$ (LBFGS) & $10^{-4}$ (manual decay) \\
optimizer & LBFGS & Adam \\
epochs & 1000 & 500\\
\multicolumn{3}{c}{ }\\
\hline
\multicolumn{3}{c}{\cellcolor{gray!20}{\textit{Parameters of the Network}}}\\
architecture & FNO & UNet (zero-padding) \\
modes / kernel size & 64 modes & $4 \times 4$ \\
depth & 6 & 4 \\
initial features &  16 & 32 \\
stride & 1 & 1 \\
dilation & 1 & 1 \\
activation function & ReLU & ReLU \\
normalization & BatchNorm & BatchNorm \\
repeat inner blocks & False & False \\
\end{tabular}\label{tab:params_fno_picnn}
\end{center}
\end{table}

\paragraph{Hyperparameters for LGCNN on random permeability}\label{app:hparams_lgcnn_randomK}
The values considered during hyperparameter optimization with Optuna~\citep{optuna_2019} and the best settings found for both steps of LGCNN are listed in~\Cref{tab:params}.  Although the optimization was originally run for 100 epochs, the optimum was consistently found within the first 25 epochs. Therefore, to reduce computation cost, we therefore conservatively lowered the maximum number of epochs to 50. This adjustment does not affect any of the reported metrics in the paper.

\begin{table}[!htbp]
\caption{LGCNN-Random $k$: Hyperparameter optimization: Parameter ranges and best configurations.}
\begin{center}
\begin{tabular}{cccc}
    \hline
    \textbf{Parameter} & \textbf{Range} & \textbf{Step 1 ($v$)} & \textbf{Step 3 ($T$)}\\
    \hline
    \hline
\multicolumn{4}{c}{ }\\
\multicolumn{4}{c}{\cellcolor{gray!20}{\textit{Parameters of the Dataset}}}\\
\multirow{2}{*}{inputs} & $v:$ $(p,i,k)$ & \multirow{2}{*}{$pik$} & \multirow{2}{*}{$iv_xv_yss_ok$} \\
& $T:$ $(i,v_x,v_y,s,s_o,k)$ & & \\
\hline
\multirow{2}{*}{skip per direction} & $v:$ 4, 8, 16, 32, 64 & \multirow{2}{*}{16} & \multirow{2}{*}{8} \\
& $T:$ 8, 16, 32, 64 & & \\
\hline
box length & 64, 128, 256, 512 & 256 & 256 \\
\multicolumn{4}{c}{ }\\
\multicolumn{4}{c}{\cellcolor{gray!20}{\textit{Parameters of Training}}}\\
loss function (training) & MAE, MSE & MSE & MAE \\
\hline
optimizer & Adam, SGD & Adam & Adam \\
\multicolumn{4}{c}{ }\\
\multicolumn{4}{c}{\cellcolor{gray!20}{\textit{Parameters of the Network}}}\\
No. initial features & 8, 16, 32, 64, 128 & 32 & 32 \\
\hline
kernel size & 3, 4, 5 & 5 & 4 \\
\hline
depth & 1, 2, 3, 4 & 4 & 4 \\
\end{tabular}
\label{tab:params}
\end{center}
\end{table}
Fixed parameters for this hyperparameter search are the learning rate (fixed at $10^{-4}$), ReLU as activation function, the batch size of 20, and the use of a batch normalization layer within the inner blocks of the UNet architecture. The validation loss used for selecting the optimal model is the MAE.

\paragraph{Hyperparameters for LGCNN on real permeability fields}\label{app:hparams_lgcnn_realK}

The values considered during hyperparameter optimization on the dataset with a more realistic permeability field were selected using Optuna~\citep{optuna_2019}, and are summarized in ~\Cref{tab:params_lgcnn_real}, along with the best configurations found for both steps of the LGCNN. The optimization was run for up to 100 epochs. For more background on the network architecture and the various hyperparameters, cf.~\ref{app:unet_block}.
Several hyperparameters were fixed during this process. These include a constant learning rate schedule, an Adam optimizer with a weight decay of $10^{-4}$, and, when enabled, a switch to LBFGS after 90 epochs. Fixed architectural parameters include a convolutional stride and dilation of 1. During training, the inputs were cut out from the full datapoints.
For model comparison, the validation loss was consistently computed using the Huber loss. 

\begin{table}[!htbp]
\caption{LGCNN-Real $k$: Hyperparameter optimization: Parameter ranges and best configurations.}
\begin{center}
    \resizebox{0.7\linewidth}{!}{
\begin{tabular}{cccc}
    \hline
    \textbf{Parameter} & \textbf{Range} & \textbf{model $\vec v$} & \textbf{model $T$} \\
    \hline
\multicolumn{4}{c}{ }\\
\multicolumn{4}{c}{\cellcolor{gray!20}{\textit{Parameters of the Dataset}}}\\
\multirow{2}{*}{inputs} & $v:$ $\nabla py, p, k, i$ & $pik$ & \\
    & $T:$ $iv_xv_yss_ok$ & &  $iv_xv_yss_ok$\\
\hline
batch size & 2, 4, 8, 16 & 8 & 8 \\
\hline
skip per direction & 256, 128, 64, 32, 16, 8 & 8 & 8 \\
\hline
box length & 1280, 640 & 1280 & 1280 \\
\multicolumn{4}{c}{ }\\
\multicolumn{4}{c}{\cellcolor{gray!20}{\textit{Parameters of Training}}}\\
loss function (training) & MSE, MAE & MSE & MSE \\
\hline
optimizer switch & True, False & False & False \\
\hline
learning rate & 1e-3, 5e-4, 1e-4, 5e-5 & 1e-4 & 1e-4 \\
\multicolumn{4}{c}{ }\\
\multicolumn{4}{c}{\cellcolor{gray!20}{\textit{Parameters of the Network}}}\\
No. initial features & 8, 16, 32 & 16 & 16 \\
\hline
kernel size & 3, 5 & 5 & 5 \\
\hline
depth & 4, 5, 6 & 6 & 6 \\
\hline
repeat inner & True, False & False & False \\
\hline
activation function & relu, tanh, sigmoid, leakyrelu & relu & relu \\
\hline
layer norm & batch-, group-, None & batch- & batch- \\
\end{tabular}
}%
\label{tab:params_lgcnn_real}
\end{center}
\end{table}

\subsection{Additional Experimental Results}\label{app:additional_experimental_results}
This section provides additional experimental results. While the main results section focused only on the test and scaling datasets, we also include here the metric values on training and validation datasets. For easier comparison, the test and scaling metrics are re-listed as well.

\paragraph{Vanilla UNet and DDUNet on \textit{101dp}: Metrics of training and ablation study}
We present metrics for predicting the temperature field directly from the inputs $pki$ using a data-driven approach, evaluated on the training, validation, and test datasets. The results are provided for two models: (1) UNet trained on 73 datapoints (73-18-10 train-validation-test split), and (2) DDUNet trained on the same 73 datapoints dataset, operating on $2\times2$ subdomains.
To assess the model's sensitivity to random initialization, the training of the same architecture was repeated five times. Based on these repetitions, the mean and standard deviation of the performance metrics were computed using the following equations:
\[
\bar{x} = \frac{1}{n} \sum_{i=1}^n x_i
\quad \text{and} \quad
\sigma = \sqrt{\frac{1}{n-1} \sum_{i=1}^n (x_i - \bar{x})^2}
\]
where \( x_i \) denotes the metric value from the \( i \)-th training run, and \( n = 5 \) is the number of runs. These results are summarized in~\Cref{tab:modelT_metrics_datadriven_pm}. The choice of $n=5$ was made empirically to balance computational effort and statistical reliability. The standard deviations in ~\Cref{tab:modelT_metrics_datadriven_pm} were used as validation: they are neither excessively large (indicating instability) nor unrealistically small (indicating insufficient sampling).

In addition to testing the UNet and DDUNet trained on 73 datapoints on the \textit{101dp} test dataset, we also evaluate these models on the same datapoint used to test the UNet$_{3dp}$. 
\begin{table}[!htbp] 
    \caption{
    Statistical error metrics for end-to-end prediction of $T$ with \textit{101dp} UNet and DDUNet. Errors in [\degree C], MSE in [\degree C$^2$], and SSIM unitless. The LGCNN-test dataset corresponds to the 1 datapoint used for testing the LGCNN approach. Mean $\pm$ standard deviation reported.}
    \centering
    \scriptsize
    \resizebox{1.0\linewidth}{!}{
    \begin{tabular}{cllc ccc}
    \hline
    \textbf{Model} & \textbf{Data} & \textbf{Case} 
    & \textbf{Huber} & \textbf{L$_{\infty}$} & \textbf{MAE} 
    & \textbf{MSE} \\
    \hline
    \hline
    \multirow{5}{*}{\rotatebox[origin=c]{90}{UNet}} 
    & \multirow{3}{*}{\rotatebox[origin=c]{90}{\shortstack{\textit{101dp}}}} 
    & train       & $0.0011 \pm 0.0002$ & $4.37 \pm 0.20$ & $0.0182 \pm 0.0020$ & $0.0023 \pm 0.0004$ \\
    & & val       & $0.0055 \pm 0.0003$ & $4.36 \pm 0.16$ & $0.0454 \pm 0.0019$ & $0.0114 \pm 0.0006$  \\
    & & test      & $0.0052 \pm 0.0003$ & $4.35 \pm 0.23$ & $0.0441 \pm 0.0019$ & $0.0110 \pm 0.0006$  \\
    & * & LGCNN-test & $0.0049 \pm 0.0002$ & $4.30 \pm 0.16$ & $0.0470 \pm 0.0010$ & $0.0102 \pm 0.0004$ \\
    & * & scaling   & $0.0017 \pm 0.0001$ & $4.48 \pm 0.15$ & $0.0208 \pm 0.0014$ & $0.0035 \pm 0.0002$ \\
    \hline
    \multirow{5}{*}{\rotatebox[origin=c]{90}{DDUNet}} 
    & \multirow{3}{*}{\rotatebox[origin=c]{90}{\shortstack{\textit{101dp}}}} 
    & train   & $0.0014 \pm 0.0003$ & $4.11 \pm 0.25$ & $0.0203 \pm 0.0026$ & $0.00300 \pm 0.0007$ \\
    & & val   & $0.0079 \pm 0.0002$ & $4.20 \pm 0.25$ & $0.0564 \pm 0.0008$ & $0.01648 \pm 0.0005$ \\
    & & test  & $0.0075 \pm 0.0001$ & $4.05 \pm 0.22$ & $0.0552 \pm 0.0008$ & $0.01580 \pm 0.0002$ \\
    & * & LGCNN-test  & $0.0057 \pm 0.0003$ & $4.00 \pm 0.20$ & $0.0526 \pm 0.0015$ & $0.01171 \pm 0.0006$ \\
    & * & scaling     & $0.0025 \pm 0.0001$ & $4.04 \pm 0.20$ & $0.0251 \pm 0.0007$ & $0.00514 \pm 0.0002$ \\
    \hline
    \end{tabular}
    }
    \label{tab:modelT_metrics_datadriven_pm}
\end{table}

\paragraph{Vanilla UNet on \textit{3dp}: Metrics of training and ablation study}
We present metrics for predicting $T$ directly from the inputs $pki$ with UNet$_{3dp}$ in \Cref{tab:model_metrics_pm}, evaluated on training, validation, test and scaling datapoints.
To assess the model's sensitivity to random initialization, the training of the same architecture was repeated five times. Other models were not trained again to reduce computational effort and because the results of all vanilla approaches indicate that the performance of UNet$_{101dp}$, UNet$_{3dp}$, and DDUNet$_{101dp}$ can be viewed as a representative of the general performance of purely data-driven methods.

\begin{table}[!htbp]
\caption{Statistical error metrics for predicting with different models and datasets. Errors in [\degree C] or [m/y], MoC in [\%], and SSIM unitless. Mean $\pm$ standard deviation reported. }
\centering
\scriptsize
\begin{tabular}{lllllrrrrr}
\toprule
\textbf{Architecture} & \textbf{Data} & \textbf{Split} & \textbf{Scaling} & \textbf{Step} & \textbf{MAE} & \textbf{RMSE} & \textbf{Huber} & \textbf{SSIM} & \textbf{MoC} \\
\midrule
UNet$^{\mathrm{a}}$ & synth.: 3dp & train & No & end-to-end & 0.0155 $\pm$ 0.0064 & 0.0732 $\pm$ 0.0310 & 0.0025 $\pm$ 0.0021 & 0.9645 $\pm$ 0.0157 & 1.6770 $\pm$ 1.9093 \\
UNet$^{\mathrm{a}}$  & synth.: 3dp & val & No & end-to-end & 0.1114 $\pm$ 0.0021 & 0.2177 $\pm$ 0.0039 & 0.0219 $\pm$ 0.0007 & 0.6251 $\pm$ 0.0067 & 6.9180 $\pm$ 0.7236 \\
UNet$^{\mathrm{a}}$ & synth.: 3dp & test & No & end-to-end & 0.1043 $\pm$ 0.0016 & 0.1917 $\pm$ 0.0035 & 0.0176 $\pm$ 0.0006 & 0.6239 $\pm$ 0.0048 & 7.7775 $\pm$ 1.5033 \\
\hline
LGCNN$^{\mathrm{b}}$ & synth.: 3dp & train & No & end-to-end & 0.0720 $\pm$ 0.0012 & 0.1736 $\pm$ 0.0019 & 0.0137 $\pm$ 0.0003 & 0.7993 $\pm$ 0.0044 & 0.0933 $\pm$ 0.0430 \\
LGCNN$^{\mathrm{b}}$ & synth.: 3dp & val & No & end-to-end & 0.1026 $\pm$ 0.0014 & 0.2095 $\pm$ 0.0032 & 0.0204 $\pm$ 0.0006 & 0.6812 $\pm$ 0.0061 & 1.6610 $\pm$ 0.0797 \\
LGCNN$^{\mathrm{b}}$ & synth.: 3dp & test & No & end-to-end & 0.0965 $\pm$ 0.0013 & 0.1866 $\pm$ 0.0030 & 0.0167 $\pm$ 0.0005 & 0.6724 $\pm$ 0.0059 & 0.1443 $\pm$ 0.0388 \\
LGCNN$^{\mathrm{b}}$ & synth.: 3dp & test & Yes & end-to-end & 0.0430 $\pm$ 0.0007 & 0.1249 $\pm$ 0.0016 & 0.0072 $\pm$ 0.0002 & 0.7908 $\pm$ 0.0151 & 0.0951 $\pm$ 0.0317 \\
UNet & synth.: 3dp & train & No & 3 & 0.0354 $\pm$ 0.0007 & 0.0770 $\pm$ 0.0024 & 0.0029 $\pm$ 0.0002 & 0.9084 $\pm$ 0.0032 & 0.1516 $\pm$ 0.0203 \\
UNet & synth.: 3dp & val & No & 3 & 0.0491 $\pm$ 0.0005 & 0.0916 $\pm$ 0.0008 & 0.0042 $\pm$ 0.0001 & 0.8251 $\pm$ 0.0033 & 0.1970 $\pm$ 0.0381 \\
UNet & synth.: 3dp & test & No & 3 & 0.0452 $\pm$ 0.0005 & 0.0829 $\pm$ 0.0011 & 0.0034 $\pm$ 0.0001 & 0.8381 $\pm$ 0.0040 & 0.1938 $\pm$ 0.0907 \\
UNet & synth.: 3dp & test & Yes & 3 & 0.0179 $\pm$ 0.0005 & 0.0399 $\pm$ 0.0007 & 0.0008 $\pm$ 0.0000 & 0.8809 $\pm$ 0.0138 & 0.1148 $\pm$ 0.0497 \\
UNet & synth.: 3dp & train & No & 1 (vx) & 14.6004 $\pm$ 10.3108 & 16.7442 $\pm$ 9.9912 & 14.1096 $\pm$ 10.3058 & 0.9884 $\pm$ 0.0020 & - \\
UNet & synth.: 3dp & train & No & 1 (vy) & 14.7965 $\pm$ 6.3568 & 17.9164 $\pm$ 6.3728 & 14.3037 $\pm$ 6.3526 & 0.9868 $\pm$ 0.0051 & - \\
UNet & synth.: 3dp & val & No & 1 (vx) & 28.7237 $\pm$ 6.9461 & 39.5616 $\pm$ 5.9981 & 28.2283 $\pm$ 6.9444 & 0.9734 $\pm$ 0.0025 & - \\
UNet & synth.: 3dp & val & No & 1 (vy) & 27.8241 $\pm$ 2.0130 & 39.7497 $\pm$ 2.7446 & 27.3289 $\pm$ 2.0125 & 0.9743 $\pm$ 0.0020 & - \\
UNet & synth.: 3dp & test & No & 1 (vx) & 27.4355 $\pm$ 5.3603 & 36.6786 $\pm$ 4.8168 & 26.9402 $\pm$ 5.3588 & 0.9727 $\pm$ 0.0025 & - \\
UNet & synth.: 3dp & test & No & 1 (vy) & 30.8755 $\pm$ 5.5264 & 42.5114 $\pm$ 6.4231 & 30.3799 $\pm$ 5.5256 & 0.9734 $\pm$ 0.0042 & - \\
UNet & synth.: 3dp & test & Yes & 1 (vx) & 28.3820 $\pm$ 5.8382 & 38.2430 $\pm$ 4.8683 & 27.8864 $\pm$ 5.8367 & 0.9770 $\pm$ 0.0017 & - \\
UNet & synth.: 3dp & test & Yes & 1 (vy) & 28.0927 $\pm$ 2.1725 & 39.6320 $\pm$ 2.8592 & 27.5972 $\pm$ 2.1719 & 0.9717 $\pm$ 0.0025 & - \\
\hline
LGCNN$^{\mathrm{b}}$ & real: 4dp & train & No & end-to-end & 0.0550 $\pm$ 0.0025 & 0.0960 $\pm$ 0.0014 & 0.0046 $\pm$ 0.0001 & 0.6744 $\pm$ 0.0249 & 1.8951 $\pm$ 0.6214 \\
LGCNN$^{\mathrm{b}}$ & real: 4dp & val & No & end-to-end & 0.0827 $\pm$ 0.0029 & 0.1553 $\pm$ 0.0031 & 0.0120 $\pm$ 0.0005 & 0.6337 $\pm$ 0.0195 & 2.0679 $\pm$ 0.4008 \\
LGCNN$^{\mathrm{b}}$ & real: 4dp & test & Yes & end-to-end & 0.0486 $\pm$ 0.0075 & 0.0714 $\pm$ 0.0056 & 0.0026 $\pm$ 0.0004 & 0.6257 $\pm$ 0.0396 & 2.7049 $\pm$ 0.6239 \\
UNet & real: 4dp & train & No & 3 & 0.0213 $\pm$ 0.0051 & 0.0295 $\pm$ 0.0061 & 0.0005 $\pm$ 0.0002 & 0.8895 $\pm$ 0.0208 & 1.9966 $\pm$ 0.2618 \\
UNet & real: 4dp & val & No & 3 & 0.0239 $\pm$ 0.0023 & 0.0395 $\pm$ 0.0034 & 0.0008 $\pm$ 0.0001 & 0.8982 $\pm$ 0.0169 & 2.9922 $\pm$ 0.3038 \\
UNet & real: 4dp & test & Yes & 3 & 0.0218 $\pm$ 0.0042 & 0.0326 $\pm$ 0.0044 & 0.0005 $\pm$ 0.0001 & 0.8734 $\pm$ 0.0190 & 1.6682 $\pm$ 0.2054 \\
UNet & real: 4dp & train & No & 1 (vx) & 22.0019 $\pm$ 5.7393 & 27.7180 $\pm$ 6.8232 & 21.5061 $\pm$ 5.7380 & 0.9731 $\pm$ 0.0280 & - \\
UNet & real: 4dp & train & No & 1 (vy) & 13.4570 $\pm$ 9.2203 & 16.8320 $\pm$ 10.8123 & 12.9699 $\pm$ 9.2128 & 0.9607 $\pm$ 0.0629 & - \\
UNet & real: 4dp & val & No & 1 (vx) & 24.8187 $\pm$ 4.0315 & 29.6583 $\pm$ 4.7384 & 24.3217 $\pm$ 4.0310 & 0.9641 $\pm$ 0.0220 & - \\
UNet & real: 4dp & val & No & 1 (vy) & 12.9956 $\pm$ 9.9800 & 15.4509 $\pm$ 10.3446 & 12.5082 $\pm$ 9.9727 & 0.9485 $\pm$ 0.0748 & - \\
UNet & real: 4dp & test & Yes & 1 (vx) & 122.7812 $\pm$ 32.6445 & 135.7671 $\pm$ 36.8100 & 122.2815 $\pm$ 32.6443 & 0.9128 $\pm$ 0.0511 & - \\
UNet & real: 4dp & test & Yes & 1 (vy) & 25.5754 $\pm$ 6.2873 & 34.6947 $\pm$ 8.2358 & 25.0801 $\pm$ 6.2858 & 0.9445 $\pm$ 0.0834 & - \\
\bottomrule
\multicolumn{10}{l}{$^{\mathrm{a}}$ newly trained with other validation loss for comparability in this table}\\
\multicolumn{10}{l}{$^{\mathrm{b}}$ with fixed first step predictions}\\
\end{tabular}\label{tab:model_metrics_pm}
\end{table}
\paragraph{LGCNN: Metrics of training and ablation study}\label{app:metrics_lgcnn}
The results of all steps/end-to-end, evaluated on both synthetic and realistic permeability fields, for the training, validation, testing, and scaling datasets are shown in \Cref{tab:model_metrics_full}.

\begin{table}[!htbp] 
\caption{Performance metrics for predicting with different models and datasets. Errors in [\degree C] or [m/y], MoC in [\%], and SSIM unitless.}
\centering
\scriptsize
\begin{tabular}{lllllrrrrr}
\toprule
\textbf{Architecture} & \textbf{Data} & \textbf{Split} & \textbf{Scaling} & \textbf{Step} & \textbf{MAE} & \textbf{RMSE} & \textbf{Huber} & \textbf{SSIM} & \textbf{MoC} \\
\midrule
UNet & synthetic: 3dp & train & No & end-to-end & 0.1181 & 0.5537 & 0.0558 & 0.8454 & 2.29 \\
UNet & synthetic: 3dp & val &  No & end-to-end & 0.1930 & 0.5795 & 0.0742 & 0.6515 & 11.32 \\
UNet & synthetic: 3dp & test & No & end-to-end & 0.1908 & 0.6190 & 0.0736 & 0.6671 & 10.85 \\
DilatedCNN & synthetic: 3dp & train & No & end-to-end & 0.1180 & 0.2494 & 0.0270 & 0.5483 & 11.28 \\
DilatedCNN & synthetic: 3dp  & val &  No &  end-to-end & 0.1273 & 0.2513 & 0.0287 & 0.5183 & 10.67 \\
DilatedCNN & synthetic: 3dp  & test & No &  end-to-end & 0.1200 & 0.2236 & 0.0237 & 0.5004 & 10.95 \\
PINN (overfit) & synthetic: 1dp & train &  No & 3 & 0.7532 & 0.8593 & 0.3454 & 0.1793 & 0 \\
\hline
LGCNN & synthetic: 3dp & train &  No &  end-to-end & 0.0657 & 0.1640 & 0.0121 & 0.8209 & 0.08 \\
LGCNN & synthetic: 3dp  & val &  No &  end-to-end & 0.0965 & 0.1996 & 0.0184 & 0.6933 & 1.54 \\
LGCNN & synthetic: 3dp  & test  & No & end-to-end & 0.0916 & 0.1738 & 0.0146 & 0.6841 & 0.09 \\
LGCNN & synthetic: 3dp  & test & Yes & end-to-end & 0.0413 & 0.1189 & 0.0065 & 0.7911 & 0.06 \\
UNet & synthetic: 3dp  & train & No  & 1 (vx) & 9.9620 & 13.0768 & 9.4732 & 0.9868 & - \\
UNet & synthetic: 3dp & val & No & 1 (vx) & 22.7179 & 33.2020 & 22.2241 & 0.9748 & - \\
UNet & synthetic: 3dp  & test  & No  & 1 (vx) & 22.3178 & 31.1860 & 21.8237 & 0.9739 & - \\
UNet & synthetic: 3dp  & test & Yes  & 1 (vx) & 24.9261 & 34.7004 & 24.4314 & 0.9784 & - \\
UNet & synthetic: 3dp & train & No & 1 (vy) & 11.8902 & 16.6018 & 11.4005 & 0.9890 & - \\
UNet & synthetic: 3dp  & val & No &  1 (vy) & 27.1026 & 39.0450 & 26.6078 & 0.9770 & - \\
UNet & synthetic: 3dp  & test  & No  & 1 (vy) & 32.7444 & 45.0703 & 32.2488 & 0.9733 & - \\
UNet & synthetic: 3dp  & test & Yes & 1 (vy) & 26.2795 & 38.2599 & 25.7847 & 0.9761 & - \\
UNet & synthetic: 3dp & train  & No &  3 & 0.0282 & 0.0632 & 0.0020 & 0.9376 & 0.11 \\
UNet & synthetic: 3dp  & val & No &  3 & 0.0452 & 0.0869 & 0.0038 & 0.8454 & 0.11 \\
UNet & synthetic: 3dp  & test  & No &  3 & 0.0417 & 0.0762 & 0.0029 & 0.8540 & 0.07 \\
UNet & synthetic: 3dp  & test & Yes &  3 & 0.0168 & 0.0373 & 0.0007 & 0.8895 & 0.08 \\
\hline
UNet, in sequence & synthetic: 3dp & train & No & 3 & 0.0402 & 0.1039 & 0.0052 & 0.9019 & 0.08 \\
UNet, in sequence & synthetic: 3dp & val &  No &  3 & 0.1013 & 0.2025 & 0.0191 & 0.6682 & 1.15 \\
UNet, in sequence & synthetic: 3dp & test & No &  3 & 0.0919 & 0.1695 & 0.0139 & 0.6714 & 1.30 \\
UNet, with zero-padding & synthetic: 3dp & train & No & 3 & 0.0440 & 0.0788 & 0.0031 & 0.8104 & 0.44 \\
UNet, with zero-padding & synthetic: 3dp & val & No  & 3 & 0.0548 & 0.0908 & 0.0041 & 0.7546 & 0.70 \\
UNet, with zero-padding & synthetic: 3dp & test & No & 3 & 0.0487 & 0.0795 & 0.0031 & 0.7784 & 0.57 \\
UNet, no partitioning & synthetic: 3dp & train & No & 1 (vx) & 3.6714 & 4.6374 & 3.1983 & 0.9959 & - \\
UNet, no partitioning & synthetic: 3dp & val &  No & 1 (vx) & 25.7367 & 37.7048 & 25.2418 & 0.9764 & - \\
UNet, no partitioning & synthetic: 3dp & test & No & 1 (vx) & 23.8584 & 33.8297 & 23.3646 & 0.9765 & - \\
UNet, no partitioning & synthetic: 3dp & train & No & 1 (vy) & 3.5177 & 4.6358 & 3.0496 & 0.9971 & - \\
UNet, no partitioning & synthetic: 3dp & val &  No & 1 (vy) & 30.1247 & 42.7471 & 29.6294 & 0.9762 & - \\
UNet, no partitioning & synthetic: 3dp & test & No & 1 (vy) & 35.9249 & 49.9040 & 35.4289 & 0.9721 & - \\
UNet, no partitioning & synthetic: 3dp & train & No & 3 & 0.0323 & 0.0736 & 0.0027 & 0.9245 & 0.14 \\
UNet, no partitioning & synthetic: 3dp & val &  No  & 3 & 0.0449 & 0.0861 & 0.0037 & 0.8333 & 0.12 \\
UNet, no partitioning & synthetic: 3dp & test & No & 3 & 0.0429 & 0.0793 & 0.0031 & 0.8349 & 0.29 \\
\hline
UNet & real: 4dp &  train & No &  1 (vx) & 14.1819 & 18.4573 & 13.6890 & 0.9915 & - \\
UNet & real: 4dp & val & No &  1 (vx) & 15.4095 & 19.5032 & 14.9187 & 0.9869 & - \\
UNet & real: 4dp   & test & Yes  & 1 (vx) & 110.0078 & 116.4903 & 109.5079 & 0.9436 & - \\
UNet & real: 4dp &  train & No  & 1 (vy) & 9.5619 & 11.2571 & 9.0680 & 0.9945 & - \\
UNet & real: 4dp & val & No &  1 (vy) & 10.6605 & 12.1757 & 10.1675 & 0.9932 & - \\
UNet & real: 4dp  & test & Yes & 1 (vy) & 17.6051 & 24.8298 & 17.1118 & 0.9905 & - \\
UNet & real: 4dp  & train & No  & 3 & 0.0139 & 0.0212 & 0.0002 & 0.9361 & 1.88 \\
UNet & real: 4dp  & val &  No  & 3 & 0.0175 & 0.0319 & 0.0005 & 0.9405 & 3.30 \\
UNet & real: 4dp  & test & Yes & 3 & 0.0189 & 0.0287 & 0.0004 & 0.8942 & 1.77 \\
LGCNN & real: 4dp & train & No  & end-to-end & 0.0534 & 0.0997 & 0.0049 & 0.6900 & 2.76 \\
LGCNN & real: 4dp & val & No &  end-to-end & 0.0841 & 0.1659 & 0.0137 & 0.6511 & 1.65 \\
LGCNN & real: 4dp & test & Yes & end-to-end & 0.0394 & 0.0665 & 0.0022 & 0.6755 & 3.50 \\
\bottomrule
\end{tabular}
\label{tab:model_metrics_full}
\end{table}

\paragraph{LGCNN+random $k$: Performance of Step 1}\label{app:performance_1st}
The model generally obtains good results in~\Cref{fig:results_T}, even for cells that are far away from injection points.
\begin{figure}[!htbp]
    \centering
    \includegraphics[width=\columnwidth]{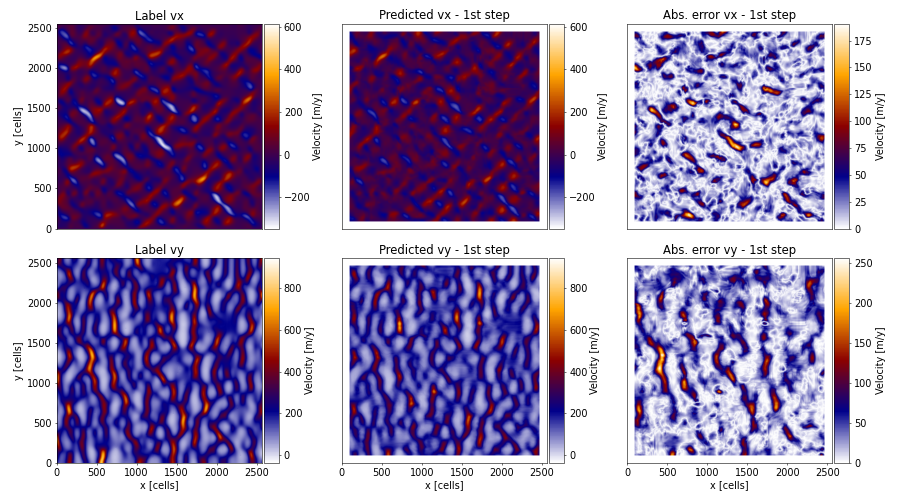}
    \caption[]
    {\small Prediction of $\Vec v$ with LGCNN. Left: Test datapoint with labels $v_x$ and $v_y$. Middle: Step 1 prediction for $v_x$, $v_y$. Right: Error of the $v_x$, $v_y$ prediction.}
    \label{fig:results_v}
\end{figure}

\begin{figure}[!htbp]
    \centering
    \includegraphics[width=0.9\columnwidth]{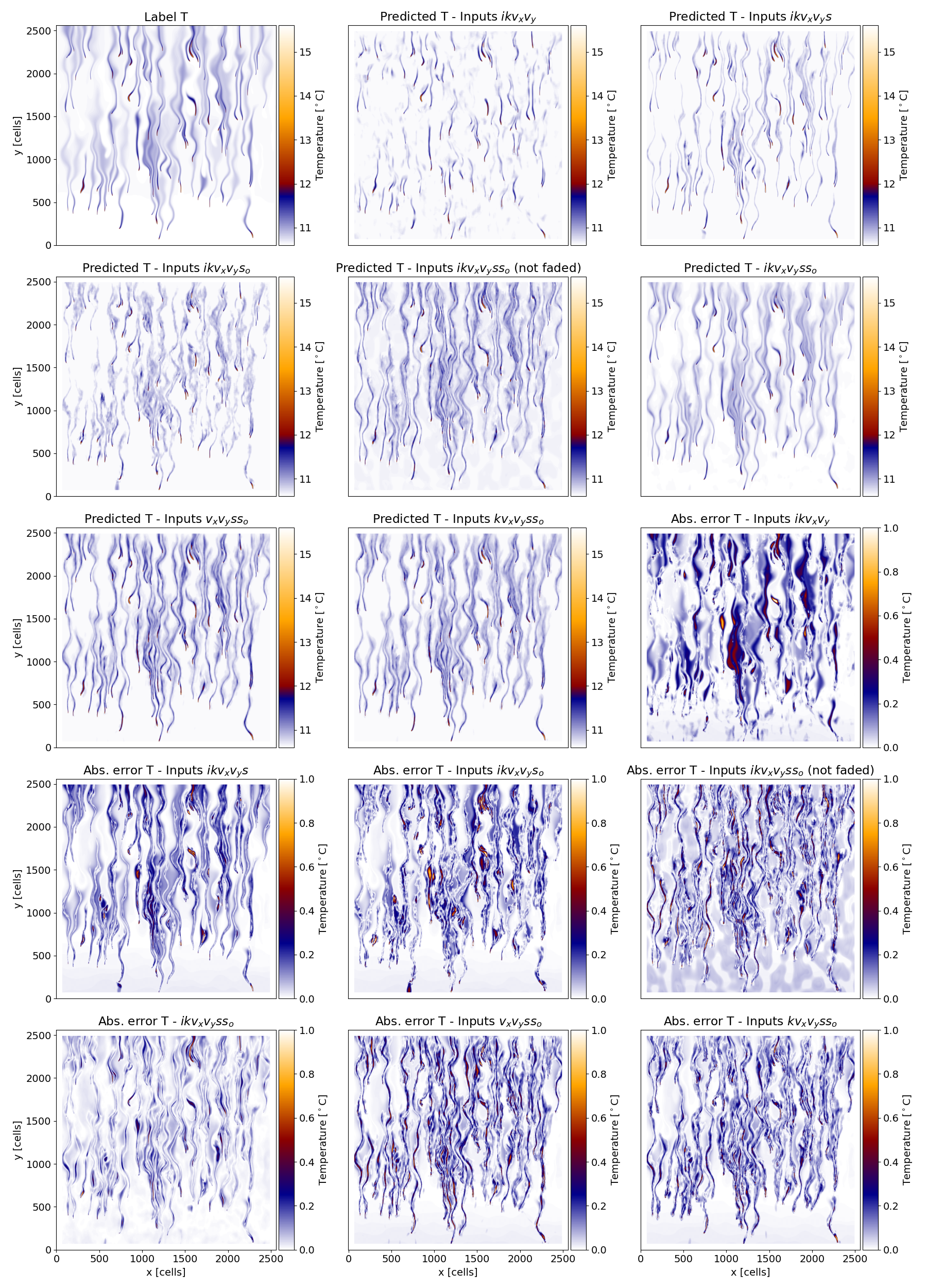}
    \caption{\small 1st Column: Label, input $k$, 2nd-5th: Without $(s,s_o)$, $(s_o)$, $(s)$, not-faded streamlines $(s,s_o)$, 6th: include all inputs, 7th-8th: Without $(i,k)$, $(i)$. Absolute errors capped at 1\degree C for better visualizations. Maximum errors are listed in \Cref{tab:exp_inputsT}.}
    \label{fig:exp_inputsT}
\end{figure}

\paragraph{LGCNN+random $k$: Experiment on Inputs to Step 3}\label{app:exp_inputs}
Additional tests show that excluding other inputs, either $i$ alone or both $i$ and $k$, raises prediction error by 58–121\%. We also evaluated alternative time-stepping schemes for solving the IVP. Replacing the 5th-order implicit Runge–Kutta method with explicit 2nd- or 4th-order schemes accelerates computation, but increases prediction error by 16–23\%—a moderate degradation compared to the complete removal of streamline inputs. Nonetheless, we retain the implicit scheme for its superior accuracy and stability. Quantitative and qualitative results for the predictions are shown in \Cref{fig:exp_inputsT} and \Cref{tab:exp_inputsT}.

\begin{table}[!htbp]
\caption{Experiment on 3rd step: Test metrics for predicting $T$ with different input combinations. Errors in [\degree C], MSE in [\degree C$^2$], and SSIM unitless.}
\centering
\scriptsize
\setlength{\tabcolsep}{2.5pt}
\begin{tabular}{lccccc}
\hline
\textbf{Inputs} & \textbf{Huber} & \textbf{L$_{\infty}$} & \textbf{MAE} & \textbf{MSE} & \textbf{SSIM} \\
\hline
\hline
$ikv_xv_y$ & 0.0070 & 2.2990 & 0.0712 & 0.0139 &  0.7662 \\
$ikv_xv_ys$ & 0.0041 & 1.8674 & 0.0545 & 0.0083 & 0.8368 \\
$ikv_xv_ys_o$ & 0.0057 & 2.6623 & 0.0598 & 0.0114 & 0.8423 \\
$ikv_xv_yss_o$ (not faded) & 0.0072 & 2.3039 & 0.0744 & 0.0144 & 0.7837 \\
$ikv_xv_yss_o^{\mathrm{a}}$ & 0.0031 & 1.8364 & 0.0442 & 0.0062 & 0.8828 \\ 
$v_xv_yss_o$ & 0.0066 & 2.1301 & 0.0647 & 0.0132 & 0.8681 \\
$kv_xv_yss_o$ & 0.0049 & 2.0925 & 0.0587 & 0.0097 & 0.8732 \\
explicit RK, order 4 & 0.0038 & 3.2636 & 0.0486 & 0.0076 & 0.8871 \\
explicit RK, order 2 & 0.0036 & 2.2034 & 0.0463 & 0.0072 & 0.8830 \\
\hline
\multicolumn{6}{l}{$^{\mathrm{a}}$ new run to be comparable to the others in this experiment: trained } \\
\multicolumn{6}{l}{with Huber validation loss, hence the results differ slightly wrt. to \Cref{tab:results_all}.}\\
\end{tabular}
\label{tab:exp_inputsT}
\end{table}

\begin{figure}[!htbp]
    \centering
    \includegraphics[width=\columnwidth]{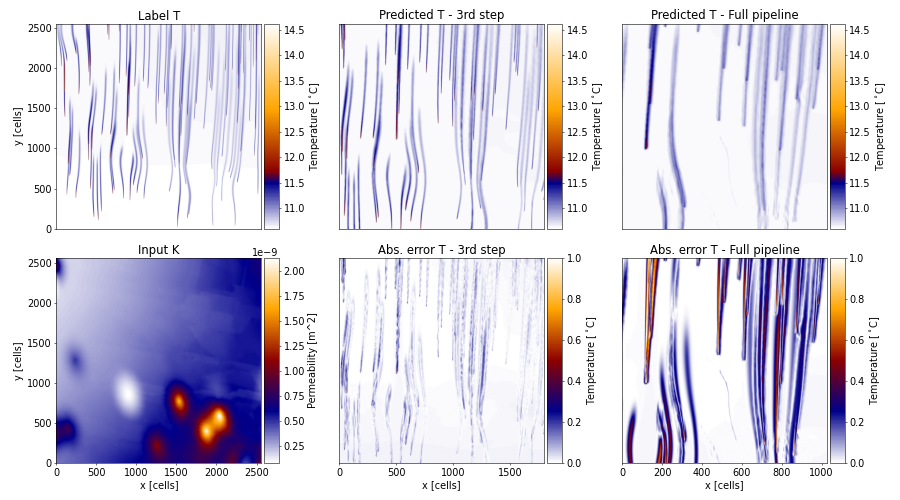}
    \caption{\small 1st Column: Label, input $k$. 2nd Column: 3rd step prediction of $T$ and error with respect to the label. 3rd Column: Predicted $T$ and error of the full pipeline.}
    \label{fig:results_T_real}
\end{figure}

\paragraph{LGCNN+realistic $k$: Performance of 3rd step and full pipeline}
The qualitative performance is observable in \Cref{fig:results_T_real}, where we see coherent streamlines and plume structures for both the isolated 3rd step and the full pipeline. 

\paragraph{Training and inference times}
\Cref{tab:times} summarizes the training and inference times, number of epochs, and dataset splits (train:val:test) for each of the three steps in our pipeline, both for the LGCNN trained on partitioned and full datasets. Although both approaches exhibit similar inference times, they show significant differences in training time, with the partitioned approach yielding better results.

\begin{table}[!htbp]
\caption{Training measurements on \textit{3dp} dataset.}
\centering
\scriptsize
\setlength{\tabcolsep}{2.5pt}
\begin{tabular}{l cccc}
\hline
\textbf{}  & \textbf{Data Points} & \textbf{Epochs$^{\mathrm{a}}$} & \textbf{Training Time$^{\mathrm{a}}$} & \textbf{Inference Time}\\
\textbf{}  & (train:val:test) & & & \\
\hline
\hline
1st Step (partitioned) & 20,736:1:1 & 19 & 9.5 min& 0.02 s\\
1st Step (full) & 1:1:1 & 9,688 & 92.6 min& 0.02 s\\
\hline
2nd Step & 1:1:1 & - & - & 9.82 s\\
\hline
3rd Step (partitioned)& 82,944:1:1 & 14 & 31.5 min & 0.03 s\\
3rd Step (full) & 1:1:1 & 9,671 & 92.1 min& 0.02 s\\
\hline
\multicolumn{5}{l}{$^{\mathrm{a}}$Early stopping: measurements until best validation loss.}\\
\end{tabular}
\label{tab:times}
\end{table}

In \Cref{tab:training_measurements}, the number of epochs and the total training time for the data-driven approaches are shown. For Step 1, both the UNet and DDUNet need many epochs and comparable training time to converge; however, for the third step and the full pipeline, the DDUNet significantly reduces both the number of epochs and the total training time required to reach convergence.

\begin{table}[!htbp]
\caption{Training measurements for the data-driven approaches trained on the \textit{101dp} dataset: UNet$_{\textit{101dp}}$ and 2$\times$2 DDUNet$_{\textit{101dp}}$.}
\centering
\begin{tabular}{l cc}
\hline
\textbf{} & \textbf{Epochs}$^{\mathrm{a}}$ & \textbf{Training Time}$^{\mathrm{a}}$ \\
\hline
\hline
\multicolumn{3}{l}{\textbf{1st Step}}\\
UNet$_{101dp}$ & 738 & 5.497 hours \\
2$\times$2 DDUNet$_{101dp}$ & 735 & 4.787 hours \\
\hline
\multicolumn{3}{l}{\textbf{3rd Step}}\\
UNet$_{101dp}$ & 726 & 8.343 hours \\
2$\times$2 DDUNet$_{101dp}$ & 303 & 3.508 hours \\
\hline
\multicolumn{3}{l}{\textbf{Full Pipeline}}\\
UNet$_{101dp}$ & 267 & 3.680 hours \\
2$\times$2 DDUNet$_{101dp}$ & 97 & 1.222 hours \\
\hline
\multicolumn{3}{l}{$^{\mathrm{a}}$Early stopping: measurements until best validation loss.}\\
\end{tabular}
\label{tab:training_measurements}
\end{table}

\subsection{Hardware Specifications}\label{app:hardware_specification}
The vanilla models that were trained on the large data-driven dataset of 101 samples were trained and evaluated on a server 
using NVIDIA V100 GPUs with 32 GB memory. All training was conducted using PyTorch 2.1.0 with CUDA 11.6 acceleration.

Training and evaluation of all other models (dilated UNet, LGCNN single steps, full pipeline and ablation studies, FNOs, PINNs) were performed on a single NVIDIA A100-SXM4 GPU. Data generation was carried out on a dual-socket system equipped with AMD EPYC 9274F CPUs.

\subsection{Glossary}\label{app:glossary}

A list of the most relevant physical properties used in our paper is provided in~\Cref{tab:glossary}.

\begin{table}[!htbp]
    \caption{Glossary of Abbreviations.}
    \begin{center}
        \begin{tabular}{r|l}
            \textbf{Abbr.}                       & \textbf{Parameter}                         \\
            \hline
            $t$                                  & time                                       \\
            $X(t_0)$                             & property X at initial time                 \\
            $X(t_{\text{end}})$                  & property X at quasi steady-state           \\
            $X_{pred}$                           & predicted property X                       \\
            $X_{sim}$                            & simulated property X                       \\
            $i$                                  & positions of heat pumps                    \\
            $Q_{inj}$                            & injected mass rate                         \\
            $\Delta T_{inj}$                     & injected temperature difference            \\
            $k$                                  & permeability                               \\
            $p$                                  & hydraulic pressure                         \\
            $g = \nabla p$                       & hydraulic pressure gradient                \\
            $\Vec v = \left ( v_x, v_y \right )$ & flow velocity                             \\
            $\Vec s$                             & both streamline fields                     \\
            $s$                                  & central streamlines, starting at all $i$          \\
            $s_o$                                & streamlines starting with $\pm$ 10 cells transversal offset compared to $s$ \\
            $T$                                  & temperature                                \\
        \end{tabular}
    \end{center}
    \label{tab:glossary}
\end{table}

\quad


\end{document}